\documentclass[11pt]{article}

\usepackage[preprint]{acl}

\usepackage{times}
\usepackage{latexsym}
\usepackage{algorithm}

\usepackage[T1]{fontenc}

\usepackage[utf8]{inputenc}

\usepackage{microtype}

\usepackage{inconsolata}

\usepackage{graphicx}
\usepackage{amsmath,amssymb}
\usepackage{array,booktabs,multirow,colortbl}
\usepackage{booktabs}
\usepackage{tabularx}
\usepackage[table]{xcolor}
\usepackage{array}
\usepackage{tabularx}
\usepackage{booktabs}
\usepackage{multirow}
\usepackage{xcolor}
\usepackage{colortbl}
\usepackage{booktabs}
\usepackage{amssymb}
\usepackage[skins,breakable]{tcolorbox}
\usepackage{xcolor}
\usepackage{colortbl}
\usepackage{enumitem}
\definecolor{rsiHeader}{HTML}{DCDCE6}
\definecolor{rsiOurs}{HTML}{EAF1E8}
\definecolor{rsiStripe}{HTML}{F2F2F2}

\definecolor{rsiModel}{HTML}{E8E8EF}

\title{RSIFlow: Adaptive Self-Improvement Workflows with an Evolving Meta-Agent}

\author{
    \textbf{Huan Zhang\textsuperscript{1,$\dagger$}},
    \textbf{Mingju Chen\textsuperscript{1,$\dagger$}},
    \textbf{Can Lv\textsuperscript{1}},
    \textbf{Dongxu Zhou\textsuperscript{1}},
    \textbf{Sen Cui\textsuperscript{3}}, \\
    \textbf{Faguo Wu\textsuperscript{1,*}},
    \textbf{Shiji Zhou\textsuperscript{1,2,*}} \\[2pt]
    {\small
    \textsuperscript{1}Beijing Advanced Innovation Center for Future Blockchain and Privacy Computing,
    School of Artificial Intelligence,} \\
    {\small
    Beihang University,
    \textsuperscript{2}Beijing Academy of Artificial Intelligence (BAAI),
    \textsuperscript{3}Tsinghua University,} \\[2pt]
    {\small
    \textsuperscript{$\dagger$}Equal contribution.
    \quad
    \textbf{Corresponding to:} Shiji Zhou
    \textless{}zhoushiji25@buaa.edu.cn\textgreater{}}
}

\begin{document}
\maketitle
\begin{abstract}
Large language model (LLM) agents have demonstrated strong capabilities in complex tasks, including code generation, tool use, and multi-hop question answering. Existing self-improvement methods enhance agent performance through model optimization, harness evolution, and executable artifact refinement, while recursive self-improvement (RSI) further extends optimization to the improvement mechanisms themselves. However, adapting self-improvement workflows to evolving performance bottlenecks while improving workflow orchestration through execution feedback remains a key challenge. We propose \textbf{RSIFlow}, a \textbf{R}ecursive \textbf{S}elf-\textbf{I}mprovement framework that integrates adaptive work\textbf{flow} orchestration with meta-agent capability evolution. RSIFlow diagnoses performance bottlenecks from task-agent trajectories and environmental feedback to dynamically select component-specific improvement workflows. The meta-agent then distills workflow outcomes into component-specific skills and cross-component principles, evolving its own harness to improve subsequent workflow orchestration. Experiments across six benchmarks and two backbone models show that RSIFlow achieves the SOTA average performance on all benchmarks. Ablation studies further demonstrate the contributions of adaptive workflow orchestration and meta-harness evolution. Code is available at https://github.com/1HUAN1/RSI-Flow.  
\end{abstract}

\begin{figure}[t]
    \centering
    \includegraphics[width=\linewidth]{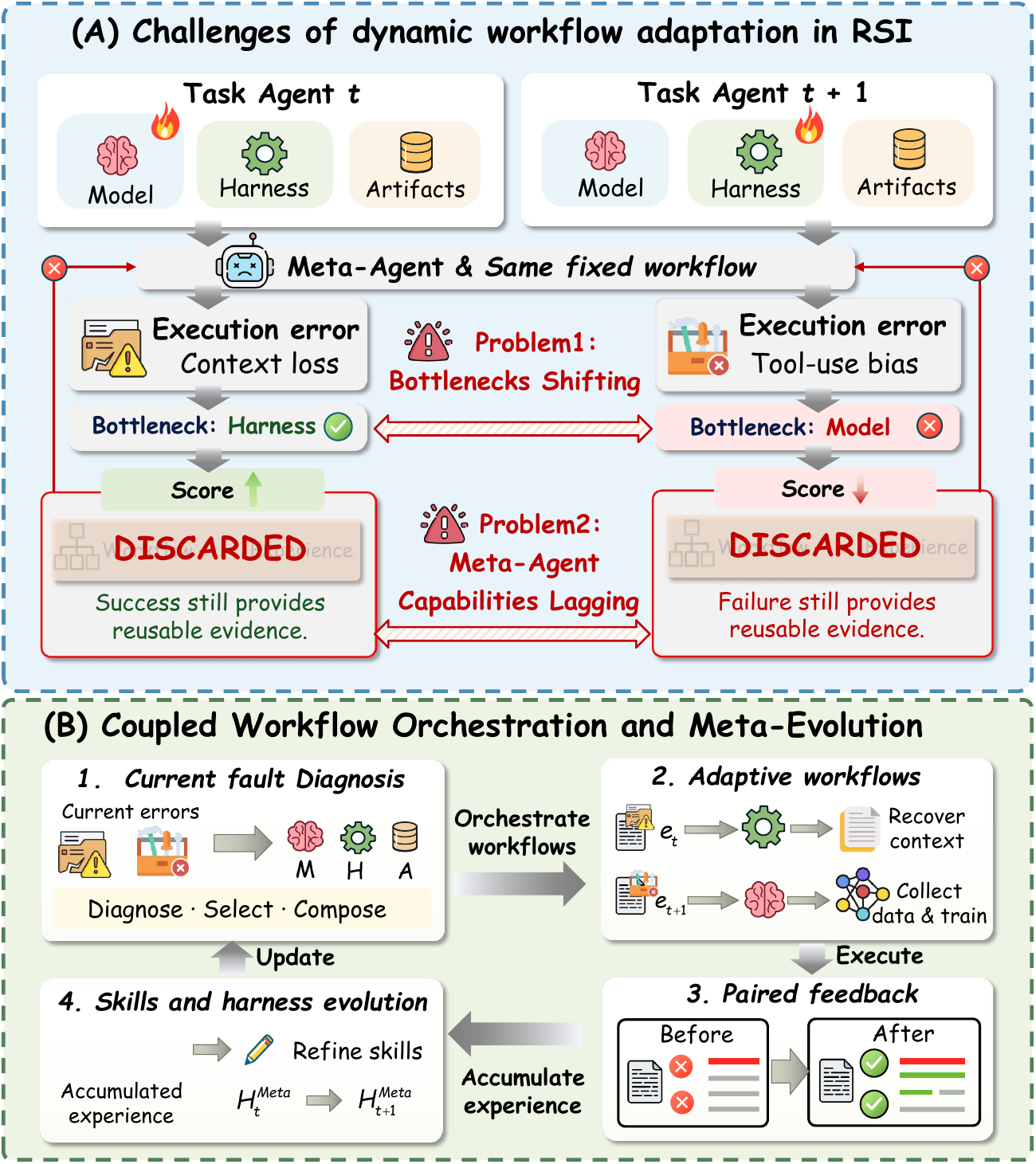}
    \caption{\textbf{Motivation of RSIFlow.}
(A) Two coupled challenges in RSI.
(B) The execution loop of RSIFlow. }
    \label{fig:intro}
    \vspace{-18 pt}
\end{figure}

\section{Introduction}
\label{sec:introduction}

Recent advances in large language models (LLMs) have enabled agents to tackle complex tasks in coding, tool use, and multi-hop question answering  \citep{NEURIPS2024_5a7c9475,NEURIPS2023_d842425e,yao2023react}. To continuously enhance the development of agents’ capabilities in task domains, existing methods have adopted complementary approaches, including  evolving execution harnesses \citep{ICLR2025_5492ecbc}, optimizing model parameters \citep{luo2025agentlightningtrainai}, and refining executable artifacts \citep{novikov2025alphaevolvecodingagentscientific}. These routes can be organized into self-improvement workflows that propose and evaluate changes using task feedback. Beyond executing such workflows, recursive self-improvement (RSI) seeks to strengthen not only task-solving capabilities but also the mechanisms that generate subsequent improvements \citep{zhang2026hyperagents,duan2026aibuilthumansgenuine}.

Research on agent self-improvement spans three increasingly broad scopes of optimization. Component-level approaches optimize a predefined part of the system, such as the model or the harness (execution workflows) \citep{luo2025agentlightningtrainai,zhang2026darwin}. System-level approaches coordinate updates across components, including joint model--harness optimization and feedback-driven selection of update targets \citep{chen2026coharnesscoevolvingharnessesmodel,chen2026harnessforgejointharnesspolicy,hebbar2026sia}. Meta-level approaches make improvement procedures editable through operator scheduling, program modification, and recursive skill evolution \citep{tan2026metarsirsi2metarecursive,zhang2026hyperagents,wang2026metaskillevolverecursiveselfimprovementllm}. These developments expand both what can be modified and how modifications are organized. However, broader editability alone does not establish how effectively a workflow addresses the agent's current bottlenecks, or whether the ability to organize that workflow improves with experience. This motivates a central question: \textit{how can RSI systems adapt self-improvement workflows to evolving bottlenecks while learning to organize these workflows more effectively?}

We formulate this challenge as \textbf{\textit{dynamic workflow adaptation}} in RSI, where adapting bottlenecks and evolving workflow orchestration must be considered together. As illustrated in Fig.~\ref{fig:intro}, self-improvement workflows must adapt to bottlenecks that vary with task requirements and the agent's capability state. Even within a fixed domain, successful updates can reshape these bottlenecks and alter the effectiveness of subsequent workflow orchestration. Meanwhile, workflow orchestration is itself a capability that can improve through experience. Feedback from workflow execution supports the accumulation and refinement of reusable knowledge for diagnosis, workflow selection, and execution. These processes form a feedback loop: task-level updates reshape improvement requirements, while workflow  experience refines the orchestration capability that guides subsequent updates.

To address this challenge, we propose \textbf{RSIFlow}, a framework that couples \textit{adaptive workflow orchestration} with \textit{meta-agent capability evolution}. We instantiate three self-improvement workflows targeting the \textit{Harness}, \textit{Model}, and \textit{Artifacts}. A meta-agent uses execution trajectories and environmental feedback to diagnose the task agent's bottlenecks, select a workflow, and orchestrate candidate generation and validation. It then distills orchestration outcomes into component-specific skills and cross-component principles to revise its own harness, including execution procedures and skill invocation. This experience-driven self-revision closes a recursive feedback loop in which workflow orchestration outcomes refine the orchestration mechanism that guides subsequent improvements.

Our main contributions are:
\begin{itemize}[leftmargin=1.35em, labelsep=0.45em, itemsep=1pt, topsep=2pt, parsep=0pt]
    \item We formulate \textit{\textbf{dynamic workflow adaptation}} as a key challenge in RSI, highlighting the coupling between evolving performance bottlenecks and continual refinement of workflow orchestration.
    \item We propose \textbf{RSIFlow}, which adaptively organizes self-improvement workflows and uses their observed outcomes to evolve the meta-agent's own harness, including its execution procedures and reusable skills.
    \item We evaluate \textbf{RSIFlow} across six benchmarks and two backbones, outperforming the strongest evaluated baseline in 11 of the 12 benchmark--backbone settings. Ablations support the benefits of both adaptive workflow orchestration and meta-harness evolution.
\end{itemize}

\section{Related Work}
\label{sec:related work}
\paragraph{Component-Specific Self-Improvement.}
Many approaches to agent self-improvement optimize a single predefined component. STaR~\citep{NEURIPS2022_639a9a17} improves model parameters by repeatedly fine-tuning on self-generated rationales that yield correct answers. AFlow~\citep{ICLR2025_5492ecbc} searches code-represented workflows, while Recursive Harness Self-Improvement~\citep{lee2026recursiveharnessselfimprovement} refines prompt-level harness specifications using pairwise feedback over revision histories. Related approaches include DSPy~\citep{ICLR2024_f1cf02ce} for language-model pipeline optimization, GEPA~\citep{ICLR2026_0e9e708b} for reflective prompt evolution, and ADAS~\citep{ICLR2025_36b7acf6} for meta-agent-guided search over agent programs. In contrast, AlphaEvolve~\citep{novikov2025alphaevolvecodingagentscientific} evolves candidate programs through execution-based evaluation, targeting task artifacts rather than the agent's improvement procedure. Self-Refine~\citep{NEURIPS2023_91edff07} iteratively revises outputs using self-feedback, while Reflexion~\citep{NEURIPS2023_1b44b878} retains verbal reflections to guide subsequent attempts and Voyager~\citep{wang2024voyager} accumulates reusable executable skills.  Co-Harness~\citep{chen2026coharnesscoevolvingharnessesmodel} and HarnessForge~\citep{chen2026harnessforgejointharnesspolicy} connect external structure and internal capability by alternating harness refinement with model optimization. Their coupled updates address model--harness compatibility, but the prescribed update order leaves open how to adapt workflow orchestration targets to evolving bottlenecks.

\begin{figure*}[t]
    \centering
    \includegraphics[width=1\textwidth]{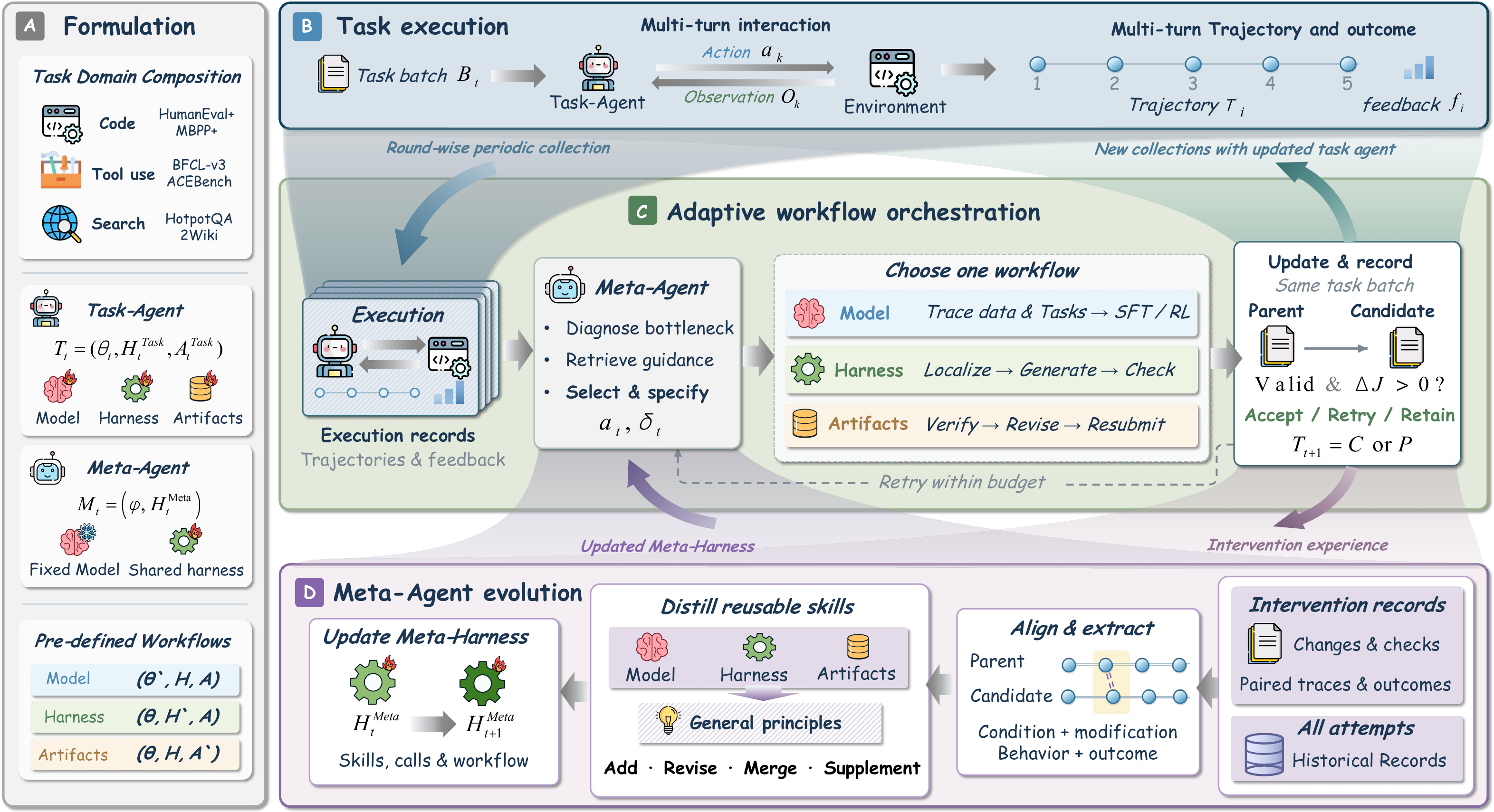}
    \caption{\textbf{Overview of RSIFlow.} \textbf{(A)} Definition of task domains and agent workflows.
\textbf{(B)} The Task-Agent executes tasks to collect trajectories and feedback.
\textbf{(C)} The Meta-Agent retrieves skills to adaptively orchestrate workflows.
\textbf{(D)} Workflow results are distilled into skills, principles and Meta-Harness.
The updated Meta-Harness guides subsequent orchestration, closing the recursive self-improvement loop.
}
\label{fig:overview}
\end{figure*}

\paragraph{Adaptive and Recursive Self-Improvement.}
Beyond prescribed alternation, SIA~\citep{hebbar2026sia} uses trajectory and performance feedback to choose between harness and weight updates, while its selection policy relies on a frozen language-model prior. HASE~\citep{luo2026harnessawareselfevolvingcoevolvingmodel} integrates task solving and harness editing into one action space and trains the policy with GRPO. A complementary direction makes the improver itself editable: Hyperagents~\citep{zhang2026hyperagents} co-evolves task-agent and meta-agent programs and demonstrates transferable improvement strategies, whereas MetaSkill-Evolve~\citep{wang2026metaskillevolverecursiveselfimprovementllm} recursively updates task skills and their improvement procedures on different timescales. Building on these complementary directions, RSIFlow couples bottleneck-aware workflow orchestration with feedback-driven Meta-Harness evolution.

\section{Methodology}
\label{sec:Methodology}

Fig.~\ref{fig:overview} provides an overview of RSIFlow. Sec.~\ref{sec:agent_formulation} formalizes the Task-Agent and Meta-Agent system and the recursive self-improvement process. Sec.~\ref{sec:adaptive_workflow_orchestration} describes how the Meta-Agent adaptively orchestrates component-specific improvement workflows. Finally, Sec.~\ref{sec:evolving meta-agent} presents how the Meta-Agent evolves through multi-level skill formation and Meta-Harness refinement.

\subsection{Preliminary}
\label{sec:agent_formulation}
\paragraph{Agent Formulation.} At evolution round $t$, we formulate the recursive self-improvement system as the joint state of a Task-Agent $T_t$ and a Meta-Agent $M_t$. The Task-Agent executes tasks, while the Meta-Agent coordinates updates to the Task-Agent's components and refines itself using feedback from updates. The system state is defined as
\begin{equation}
\begin{gathered}
X_t = (T_t,M_t),\quad M_t = (\phi,H_t^{\mathrm{Meta}}),\\
T_t = (\theta_t,H_t^{\mathrm{Task}},A_t^{\mathrm{Task}}).
\end{gathered}
\label{eq:agent_formulation}
\end{equation}
Here, $\theta_t$ denotes the Task-Agent's model parameters, $H_t^{\mathrm{Task}}$ denotes the external harness, and $A_t^{\mathrm{Task}}$ denotes task-level
deliverables, such as executable code and solutions, that can be refined
through environmental feedback. The Meta-Agent's base model parameters $\phi$ remain fixed throughout the process, and its evolution occurs entirely through modifications to its harness $H_t^{\mathrm{Meta}}$.

We define a harness as the execution interface external to the model and adopt the same three-component abstraction for both agents:
\begin{equation}
\begin{gathered}
H_t^r = \left(\mathrm{Action}_t^r,\mathrm{Skill}_t^r,\mathrm{Planning}_t^r\right),\\
r \in \{\mathrm{Task},\mathrm{Meta}\}.
\end{gathered}
\label{eq:harness_composition}
\end{equation}
\textit{Action} organizes execution and information actions. \textit{Skill} provides reusable experience and principles, while \textit{Planning} decomposes objectives into steps. The Task-Agent harness supports task execution, whereas the Meta-Agent harness guides diagnosis, component selection and improvement.

\subsection{Adaptive workflow orchestration}
\label{sec:adaptive_workflow_orchestration}

\paragraph{Workflow selection and instantiation.}
The Meta-Agent adaptively orchestrates three
component-specific improvement workflows:
\begin{equation}
a_t \in
\{\mathrm{Model},\mathrm{Harness},\mathrm{Artifacts}\}.
\label{eq:adaptive_routing}
\end{equation}
Using execution trajectories, environmental feedback, and accumulated meta-skills, the Meta-Agent selects a component to modify while preserving the other two. All candidates share the same parent $T_t$ and evaluation batch. The first valid candidate with a positive gain is accepted as $T_{t+1}$; otherwise, $T_{t+1}=T_t$. Feedback from all update attempts informs subsequent Meta-Agent evolution.

\paragraph{Model update.}
The Meta-Agent converts task experience into parameter updates.
It constructs training data $\mathcal D_t$ from execution
trajectories and environmental feedback, then optimizes the
following objective starting from $\theta_t$:
\begin{equation}
\min_{\theta}\;
\mathcal L(\theta;\mathcal D_t)
=
\mathbb E_{z\sim\mathcal D_t}
\bigl[\ell(\theta;z)\bigr].
\label{eq:model_update_loss}
\end{equation}
Here, $z$ is a training sample or sample group, and $\ell$
is the method-specific loss. Our default update uses SFT to learn from successful trajectories collected during task execution. Reusing these trajectories reduces the exploration cost of acquiring training data. The candidate inherits the updated parameters $\theta'$ alongside the parent's Harness and Artifacts. We also evaluate an RL-based model-update variant,
with details provided in App. ~\ref{app:model}.

\paragraph{Harness update.}
The Meta-Agent uses execution feedback to select a component
set $\mathcal I_t\subseteq\mathcal C$, where
$\mathcal C=\{\mathrm{Action},\mathrm{Skill},\mathrm{Planning}\}$.
It follows a localize--refine--generate--validate procedure
to produce new implementations $\widehat H_t^{\mathcal I_t}$.
The candidate Harness replaces the selected
components while preserving the others:
\begin{equation}
H'
=
H_t^{\mathrm{Task}}
\left[\mathcal I_t\leftarrow
\widehat H_t^{\mathcal I_t}\right].
\label{eq:harness_update}
\end{equation}
The replacement operator rewrites the selected components and
preserves the rest. Revisions may change prompts, context
organization, memory retrieval, planning and reasoning rules,
tool orchestration, or internal verification, subject to the
corresponding interface and permission constraints.
For example, jointly revising memory retrieval
conditions and planning rules constitutes a single
Harness update. Model parameters remain fixed.


\paragraph{Artifacts refinement.}
For each task $x \in B_t$, let $A_{t,x}$ denote
its current output. The Meta-Agent derives
diagnostic guidance $\delta_{t,x}$ from execution
trajectories $\tau_{t,x}$ and feedback $f_{t,x}$.
Keeping the Task-Agent's model and harness fixed,
candidate revisions are generated as
\begin{equation}
\hat{A}_{t,x}^{(j)} \sim
p_{\theta_t,H_t^{\mathrm{Task}}}
\left(
\cdot \mid
x,A_{t,x},\tau_{t,x},f_{t,x},\delta_{t,x}
\right),
\label{eq:artifacts_generation}
\end{equation}
where $j \leq L$ indexes bounded revision attempts.
Let $\hat{A}_t^{(j)}$ collect the revised outputs
and $\hat{T}_t^{(j)} =
(\theta_t,H_t^{\mathrm{Task}},\hat{A}_t^{(j)})$.
Each candidate is evaluated against the same
task batch:
\begin{equation}
\Delta J_t^{(j)} =
J(\hat{T}_t^{(j)};B_t)
-
J(T_t;B_t).
\label{eq:artifacts_gain}
\end{equation}
Here, \(B_t\) denotes a training batch. The gain-based acceptance rule is not applied to evaluation set. The first valid candidate yielding a positive performance gain is selected:
\begin{equation}
j^* = \min
\left\{
j \leq L \;\middle|\;
\mathcal{V}(\hat{A}_t^{(j)})=1,\;
\Delta J_t^{(j)}>0
\right\},
\label{eq:artifacts_selection}
\end{equation}
where $\mathcal{V}$ enforces output validity
and execution constraints. The Meta-Agent provides diagnostic guidance without directly modifying
task artifacts, while $\theta_t$ and $H_t^{\mathrm{Task}}$ remain unchanged. 

\paragraph{Recursive self-improvement workflow.}
At evolution round $t$, the Meta-Agent $M_t$
orchestrates and evaluates candidate updates to
the Task-Agent $T_t$, yielding the next-round
Task-Agent $T_{t+1}$ and workflow experience $e_t$.
It distills component-specific skills and general
principles from this experience, integrating them
into its Meta-Harness to form $M_{t+1}$.
The updated knowledge guides diagnosis, workflow
selection, and candidate generation in subsequent
rounds as the Task-Agent's bottlenecks evolve.
Together, $(T_{t+1},M_{t+1})$ enter the
next evolution round, allowing accumulated workflow
experience to continuously refine the orchestration
mechanism. Details are provided in
App.~\ref{app:algorithm}.

\subsection{Meta-agent Capability Evolution}
\label{sec:evolving meta-agent}

\paragraph{Difference extraction and skill formation.}
The Meta-Agent aligns parent and candidate trajectories
for the same task through shared subgoals, tool events,
or error feedback. These paired comparisons connect
candidate modifications to local behavioral changes
and task-level outcomes.
From these comparisons and workflow experience
$e_t$, the Meta-Agent distills two complementary
forms of reusable knowledge: component-specific
skills $\mathcal S_t$ for Model, Harness, and Artifacts
updates, and general principles $\mathcal P_t$ for
cross-component diagnosis and workflow selection.
Each skill or principle is represented as
\begin{equation}
k = (\Omega_k, a_k, \rho_k, \Gamma_k, \hat r_k, E_k),
\label{eq:skill_representation}
\end{equation}
where $\Omega_k$ specifies applicability conditions,
$a_k$ update scope, $\rho_k$ conditional guidance,
$\Gamma_k$ preservation requirements, $\hat r_k$
expected behavioral response, and $E_k$ supporting
and contradicting evidence.

\paragraph{Evidence-driven skill maintenance.}
The Meta-Agent integrates newly distilled skills
$\mathcal S_t$ and general principles $\mathcal P_t$
into the existing library $\mathcal K_t$ through
four evidence-driven operations.
\textit{\textbf{Add}} introduces previously uncovered
strategies with their applicability and source evidence.
\textit{\textbf{Supplement}} enriches existing entries with
new outcome evidence while preserving their guidance.
\textit{\textbf{Revise}} refines applicability conditions,
guidance, or preservation requirements through
linked successor versions.
\textit{\textbf{Merge}} consolidates overlapping or superseded
entries while preserving accumulated evidence
and provenance.
These operations update the library as
\begin{equation}
\begin{aligned}
\mathcal C_t &= \mathcal S_t \cup \mathcal P_t,\\
\mathcal K_{t+1}
&= (\mathcal K_t \setminus \mathcal R_t)
   \cup \mathcal C_t^{+}.
\end{aligned}
\label{eq:skill_library_update}
\end{equation}
Here, $\mathcal C_t^{+}$ contains records introduced
or updated through the four operations, while
$\mathcal R_t \subseteq \mathcal K_t$ denotes entries
replaced or consolidated in the active library.
The resulting library combines component-specific
skills and cross-component principles while retaining
traceable evidence and revision histories across rounds.
Further details are provided in
App. ~\ref{app:skill-maintenance}.
\paragraph{Integration into the Meta-Harness.}
The Meta-Agent operationalizes $\mathcal K_{t+1}$
by updating the Action interfaces, retrieval calls,
execution instructions, and supplementary prompts
that invoke this knowledge to guide subsequent
improvement decisions:
\begin{equation}
\begin{gathered}
H_{t+1}^{\mathrm{Meta}}
= (\mathrm{Action}_{t+1}^{\mathrm{Meta}},
   \mathcal K_{t+1},
   \mathrm{Planning}_t^{\mathrm{Meta}}),\\
M_{t+1}
= (\phi,H_{t+1}^{\mathrm{Meta}}).
\end{gathered}
\label{eq:meta_harness_edit}
\end{equation}
The evolved Meta-Harness guides subsequent bottleneck
diagnosis, workflow selection, and candidate generation.
Thus, experience accumulated from Task-Agent updates
recursively improves workflow orchestration, while
the Meta-Agent's model parameters $\phi$ remain fixed.

\begin{table*}[t]
\centering
\caption{Main results across six benchmarks with Qwen3-4B and Qwen3-8B backbones. Best results are highlighted in bold.}
\label{tab:main_comparison}
\begingroup
\setlength{\tabcolsep}{3pt}
\renewcommand{\arraystretch}{1.08}
\setlength{\aboverulesep}{1pt}
\setlength{\belowrulesep}{1pt}
\setlength{\abovetopsep}{0pt}
\setlength{\belowbottomsep}{0pt}


\begin{tabularx}{\linewidth}{l*{4}{c}*{3}{>{\centering\arraybackslash}X}}
\toprule[0.9pt]

\multirow{2}{*}{\textbf{Method}}
& \multicolumn{2}{c}{\textbf{Code}}
& \multicolumn{2}{c}{\textbf{Tool Use}}
& \multicolumn{2}{c}{\textbf{SearchQA}}
& \multirow{2}{*}{\textbf{Overall}} \\
\cmidrule[0.3pt](lr){2-3}
\cmidrule[0.3pt](lr){4-5}
\cmidrule[0.3pt](lr){6-7}

& \textbf{HumanEval+}
& \textbf{MBPP+}
& \textbf{BFCL-v3}
& \textbf{ACEBench}
& \textbf{HotpotQA}
& \textbf{2Wiki}
& \\
\midrule[0.8pt]

\addlinespace[1pt]
\multicolumn{8}{c}{\textbf{Qwen3-4B}} \\
\addlinespace[1pt]
\midrule[0.3pt]

AFlow
& 39.63 & 53.37 & 26.31 & 29.82 & 52.59 & 38.45 & 40.03\\
\rowcolor{gray!12}
MermaidFlow
& 48.78 & 57.87 & 28.19 & 31.58 & 56.32 & 38.79 & 43.59 \\
\addlinespace[1.2pt]

SFT
& 39.63 & 55.62 & 26.88 & 29.82 & 53.52 & 41.24 & 41.12 \\
\rowcolor{gray!12}
GRPO
& 43.90 & 55.06 & 22.30 & 33.33 & 54.39 & 39.32 & 41.38 \\
\addlinespace[1.2pt]

HarnessForge
& 54.27 & 62.36 & 22.39 & 25.12 & 52.38 & 30.91 & 41.24 \\
\rowcolor{gray!12}
SIA
& 53.66 & 60.11 & 29.00 & 33.91 & 52.51 & 41.68 & 45.15 \\
\addlinespace[1.2pt]

Hyperagent
& 43.90 & 52.81 & 26.88 & 31.57 & 51.82 & 40.59 & 41.26 \\
\addlinespace[1.2pt]
\rowcolor{rsiOurs}
\textbf{RSIFLow}
& \textbf{68.29} & \textbf{66.29} & \textbf{29.25} & \textbf{38.00} & \textbf{56.72} & \textbf{48.72} & \textbf{51.21} \\

\midrule[0.8pt]

\addlinespace[1pt]
\multicolumn{8}{c}{\textbf{Qwen3-8B}} \\
\addlinespace[1pt]
\midrule[0.3pt]

AFlow
& 68.29 & 69.10 & 34.89 & 51.46 & 53.55 & 48.32 & 54.27 \\
\rowcolor{gray!12}
MermaidFlow
& 69.51 & 68.54 & 30.96 & 46.20 & 56.29 & 42.70 & 52.37 \\
\addlinespace[1.2pt]

SFT
& 68.29 & 69.10 & 37.01 & 40.35 & 51.57 & 44.49 & 51.80 \\
\rowcolor{gray!12}
GRPO
& 70.12 & 69.66 & 34.31 & 38.59 & 49.81 & 46.33 & 51.47 \\
\addlinespace[1.2pt]

HarnessForge
& 70.12 & 67.42 & 38.15 & \textbf{59.07} & 45.83 & 37.12 & 52.95 \\
\rowcolor{gray!12}
SIA
& 73.78 & 70.22 & 40.36 & 51.46 & 53.96 & 46.43 & 56.04 \\
\addlinespace[1.2pt]

Hyperagent
& 71.34 & 69.10 & 34.31 & 44.44 & 53.47 & 48.73 & 53.57 \\
\addlinespace[1.2pt]
\rowcolor{rsiOurs}
\textbf{RSIFLow}
& \textbf{76.83}
& \textbf{74.16}
& \textbf{41.34}
& 52.62
& \textbf{62.67}
& \textbf{52.08}
& \textbf{59.95} \\
\bottomrule[0.9pt]
\end{tabularx}
\endgroup
\end{table*}

\section{Experiments}

\subsection{Experimental Setup}
\label{sec:experimental_setup}

\paragraph{Benchmarks.} We train \textbf{RSIFlow} on training data from three task domains and evaluate it on six public benchmarks spanning the same domains:
\textbf{(1) code generation,} HumanEval+ and MBPP+~\citep{NEURIPS2023_43e9d647}.
\textbf{(2) tool use,} BFCL-v3~\citep{pmlr-v267-patil25a} and ACEBench~\citep{chen-etal-2025-acebench}.
\textbf{(3) search-QA,} HotpotQA~\citep{yang-etal-2018-hotpotqa} and 2Wiki~\citep{ho-etal-2020-constructing}.
Detailed descriptions and evaluation metrics are in App. ~\ref{app:metrics}

\paragraph{Baselines.} We compare RSIFlow with three series of baselines:
\textbf{(1) model-only evolution}, including SFT~\citep{NEURIPS2022_b1efde53} and GRPO~\citep{shao2024deepseekmathpushinglimitsmathematical};
\textbf{(2) harness-only evolution}, including AFlow~\citep{ICLR2025_5492ecbc} and MermaidFlow~\citep{zheng2025mermaidflowredefiningagenticworkflow};
\textbf{(3) model-harness both evolution,} including HarnessForge~\citep{chen2026harnessforgejointharnesspolicy} and SIA~\citep{hebbar2026sia}, Hyperagent~\citep{zhang2026hyperagents} further evolves the harness of the meta agent, incorporating the evolution of the meta into the cycle.
Detailed configurations of baselines are provided in App. ~\ref{app:baselines}

\paragraph{Implementation Details.} We initialize the Task-Agent of \textsc{RSIFlow} with Qwen3-4B and Qwen3-8B~\citep{yang2025qwen3technicalreport} as the backbone, respectively.
The Meta-Agent uses GPT-5.6 Sol as its base model.
We run $R=5$ evolution rounds and report the main results with a maximum of two attempts per evolution round.
Additional implementation details are provided in App. ~\ref{app:algorithms}

\subsection{Main Results}
\label{sec:main_results}

Tab.~\ref{tab:main_comparison} compares \textsc{RSIFlow} against model-only, harness-only, and joint model--harness baselines on six agentic benchmarks and two backbone sizes, averaging +3.42\% over the strongest per-metric baselines. \textsc{RSIFlow} leads in 11 of 12 benchmark--backbone settings across code generation, tool use, and search-QA. Its overall scores reach 51.21\% with Qwen3-4B and 59.95\% with Qwen3-8B, surpassing the strongest baseline by 6.06\% and 3.91\%, respectively. The largest gain occurs on HumanEval+, where \textsc{RSIFlow} improves pass@1 by 14.02\% with Qwen3-4B. On MBPP+, pass@1 improves by an average of 3.94\% across backbones. For tool use, \textsc{RSIFlow} achieves the highest BFCL-v3 scores with both backbones and the best ACEBench result with Qwen3-4B. On retrieval-intensive tasks, it improves F1 by an average of 5.20\% on 2Wiki across backbones and by 6.38\% on HotpotQA with Qwen3-8B.

\begin{figure*}[t]
    \centering
    \includegraphics[width=1.0\linewidth]{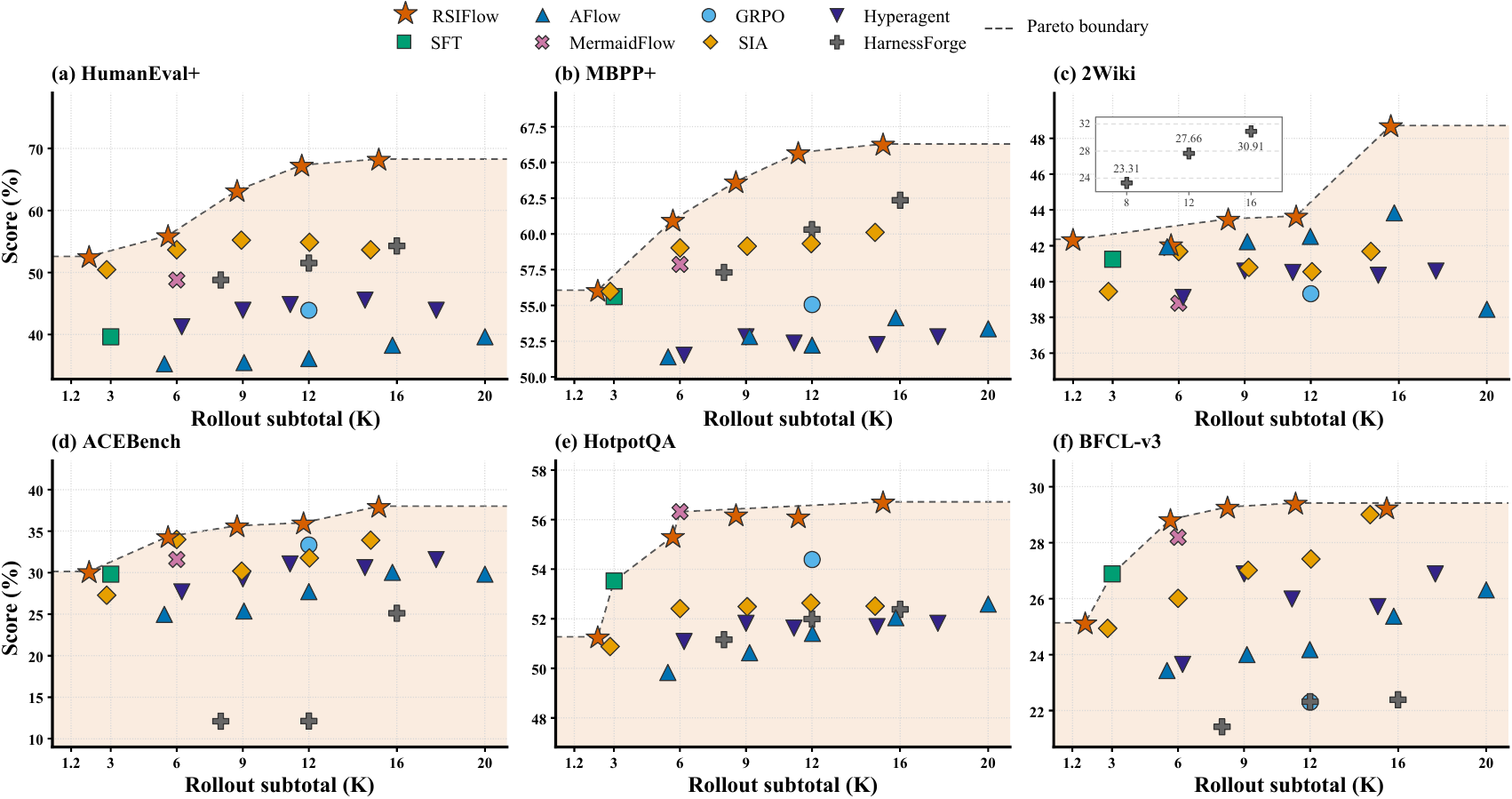}
    \caption{Budget–performance Pareto analysis across all six benchmarks with Qwen3-4B backbone. }
    \label{fig:budget-performance}
\end{figure*}

\subsection{Ablation Study}
\label{sec:ablation}

Table~\ref{tab:rsi_core_ablation} evaluates adaptive workflow orchestration and Meta-Agent evolution over five rounds. Compared with Vanilla, RSIFlow improves Code, Tool Use, and SearchQA by 19.04\%, 3.93\%, and 18.20\% , respectively, increasing the three-domain average by 13.72 \%. While fixed-component workflows exhibit domain-specific strengths, adaptive orchestration achieves the highest scores across all three domains, outperforming the fixed workflow by 4.13\% - 8.82\%  on average. Overall, RSIFlow exceeds the average of all four ablated variants by 6.18 \%. Disabling Meta-Agent evolution reduces average performance by 5.74 \% , highlighting the importance of distilling historical feedback into reusable skills and principles. These results demonstrate that adaptive component orchestration enables balanced cross-domain improvements, while experience-driven Meta-Harness evolution further strengthens workflow effectiveness.

\subsection{Framework Analysis}
\label{sec:framework_analysis}
\begin{table}[t]
\centering
\caption{Module ablation of RSIFlow using Qwen3-
4B under $R=5$ evolution rounds. With meta evolution enabled, we compare fixed task-evolution routines targeting the harness (H), model (M), or artifacts (A). }
\label{tab:rsi_core_ablation}

\begingroup
\footnotesize

\setlength{\tabcolsep}{1.5pt}
\renewcommand{\arraystretch}{1.65}

\setlength{\aboverulesep}{2.5pt}
\setlength{\belowrulesep}{2.5pt}
\setlength{\abovetopsep}{0pt}
\setlength{\belowbottomsep}{0pt}

\newcommand{\rsiGain}[2]{%
    \mbox{#1\,{\scriptsize\textcolor{green!45!black}{($\uparrow$#2)}}}%
}

\newcommand{\rsiSame}[1]{%
    \mbox{#1\,{\scriptsize\textcolor{gray}{($\rightarrow$0.00)}}}%
}

\begin{tabularx}{\columnwidth}{
    >{\hsize=0.55\hsize\linewidth=\hsize\centering\arraybackslash}X
    >{\hsize=0.70\hsize\linewidth=\hsize\centering\arraybackslash}X
    *{3}{>{\hsize=1.25\hsize\linewidth=\hsize\centering\arraybackslash}X}
}
\toprule[0.9pt]

\textbf{Meta}
& \textbf{Task}
& \multirow[c]{2}{*}{\textbf{Code}}
& \multirow[c]{2}{*}{\textbf{Tool use}}
& \multirow[c]{2}{*}{\textbf{SearchQA}} \\

\textit{Evo.}
& \textit{Rou.}
& & & \\

\midrule[0.6pt]

\rowcolor{gray!12}
\multicolumn{2}{c}{Vanilla}
& 48.25 & 29.70 & 34.52 \\

\midrule[0.3pt]

$\times$ & $\checkmark$
& \rsiGain{56.73}{8.48}
& \rsiSame{29.70}
& \rsiGain{49.98}{15.46} \\

$\checkmark$ & $\times(H)$
& \rsiGain{63.42}{15.17}
& \rsiGain{32.93}{3.23}
& \rsiGain{44.89}{10.37} \\

$\checkmark$ & $\times(M)$
& \rsiGain{50.31}{2.06}
& \rsiGain{32.15}{2.45}
& \rsiGain{44.73}{10.21} \\

$\checkmark$ & $\times(A)$
& \rsiGain{52.34}{4.09}
& \rsiGain{30.69}{0.99}
& \rsiGain{52.57}{18.05} \\

\addlinespace[3pt]

\rowcolor{rsiOurs}
$\checkmark$ & $\checkmark$
& \rsiGain{\textbf{67.29}}{\textbf{19.04}}
& \rsiGain{\textbf{33.63}}{\textbf{3.93}}
& \rsiGain{\textbf{52.72}}{\textbf{18.20}} \\

\bottomrule[0.9pt]
\end{tabularx}
\endgroup
\end{table}

\paragraph{Rollout-budget and performance efficiency.} 
\label{sec:rollout-budget and performance efficiency}
Fig.~\ref{fig:budget-performance} compares performance under varying rollout budgets on all six benchmarks. RSIFlow generally lies on or near the empirical
Pareto frontier and achieves the highest observed score on each
benchmark. This advantage is consistent with its adaptive component
selection and Task--Meta co-evolution: RSIFlow reuses rollout
evidence to guide targeted updates to the Task-Agent's Model,
Harness, or Artifacts, rather than relying solely on additional
exploration under a fixed harness. The co-evolving Meta-Agent
further distills workflow outcomes into reusable skills and
refines its improvement procedure, allowing accumulated experience
to guide subsequent updates. Together, these mechanisms help
concentrate the rollout budget on informed workflow orchestration. Overall, RSIFlow offers a stronger performance--budget trade-off than the evaluated baselines.

\paragraph{Meta-agent backbone sensitivity. }

We evaluate RSIFlow with three Meta-Agent backbones over
five evolution rounds on HumanEval+. As shown in Fig.~4,
all three models achieve overall performance improvements,
with gains of 11.59\%, 4.88\%, and 24.39\% for
Claude-Sonnet-5, DeepSeek-V4.1-Flash, and GPT-5.6-Sol,
respectively. GPT-5.6-Sol achieves the highest final
performance of 68.29\%, albeit with greater API token
consumption. From Round 3 to Round 5, it improves by
12.80\%, compared with 6.10\% and 4.27\% for the other
two backbones, suggesting that stronger Meta-Agents may
better leverage accumulated intervention feedback.
These results demonstrate that RSIFlow supports different
Meta-Agent backbones, while the Meta-Agent's underlying
capabilities influence the cumulative effectiveness of
recursive self-improvement.

\paragraph{Meta-evolution necessity analysis.}
\label{sec:meta-evolution necessity analysis.}
Table~3 compares different Meta versions improving the same initial Task-Agent, with scores averaged over two benchmarks per domain. Meta0 improves SearchQA by 10.10\%, reaching 44.62\% through harness updates. Although Meta4 also selects the harness, it further outperforms Meta0 by 20.56\% and 9.51\% on Code and SearchQA, respectively. Meta3 achieves the highest Tool Use score of 31.83\% through artifact updates. These results indicate that the evolution of meta-agent capabilities not only enables better adaptive workflow orchestration but also enhances the performance of workflow orchestration across different task domains. Further analysis is provided in App.~\ref{app:meta-compatibility}.

\begin{figure}[t]
    \centering
    \includegraphics[width=1.0\linewidth]{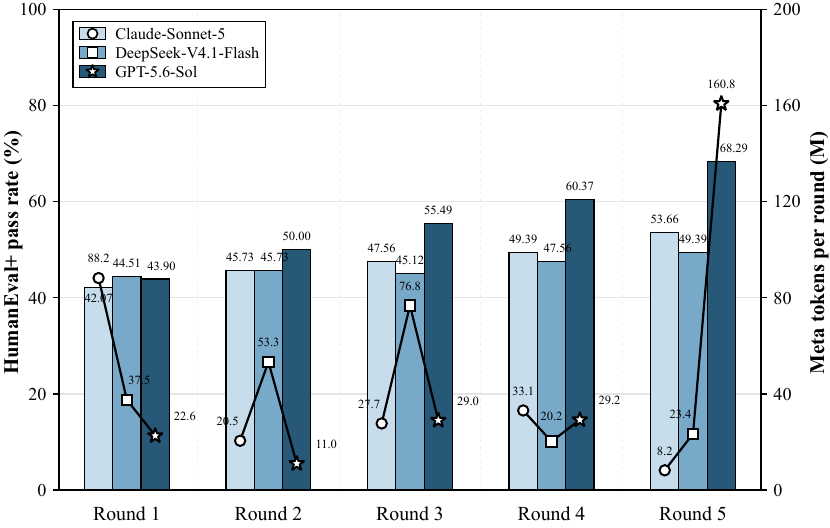}
    \caption{Sensitivity of RSIFlow to different Meta-Agent backbones on HumanEval+.}
    \label{fig:API_sensitivity_analysis}
\end{figure}

\begin{table}[t]
\centering
\caption{Domain-level performance of selected Meta candidates. Scores are macro-averages (\%) across two benchmarks per domain. Component refers to the workflow selected for orchestration by meta agent. }
\label{tab:meta_domain_averages}

\begingroup
\footnotesize

\setlength{\tabcolsep}{2.5pt}
\renewcommand{\arraystretch}{1.65}

\setlength{\aboverulesep}{2.5pt}
\setlength{\belowrulesep}{2.5pt}
\setlength{\abovetopsep}{0pt}
\setlength{\belowbottomsep}{0pt}

\vspace{3pt}

\begin{tabularx}{\columnwidth}{@{}
    >{\hsize=0.70\hsize\linewidth=\hsize
      \centering\arraybackslash}X
    >{\hsize=1.30\hsize\linewidth=\hsize
      \centering\arraybackslash}X
    *{3}{>{\centering\arraybackslash}X}
@{}}
\toprule[0.9pt]

\textbf{Meta}
& \textbf{Component}
& \textbf{Code}
& \textbf{Tool Use}
& \textbf{SearchQA} \\
\midrule[0.6pt]
\rowcolor{gray!12}
\multicolumn{2}{c}{Vanilla}
& 48.25 & 29.70 &  34.52\\
\midrule[0.3pt]

Meta0 & HARNESS
& 48.30 & 30.40 & 44.62 \\

Meta1 & HARNESS
& 46.32 & 31.16 & 45.04 \\

Meta2 & MODEL
& 67.90 & 31.51 & 43.99 \\

Meta3 & ARTIFACTS
& 67.08 & \textbf{31.83} & 50.58 \\

\rowcolor{rsiOurs}
Meta4 & HARNESS
& \textbf{68.86} & 30.84 & \textbf{54.13} \\

\bottomrule[0.9pt]
\end{tabularx}

\endgroup
\end{table}

\begin{figure*}[t]
    \centering
    \includegraphics[width=1\textwidth]{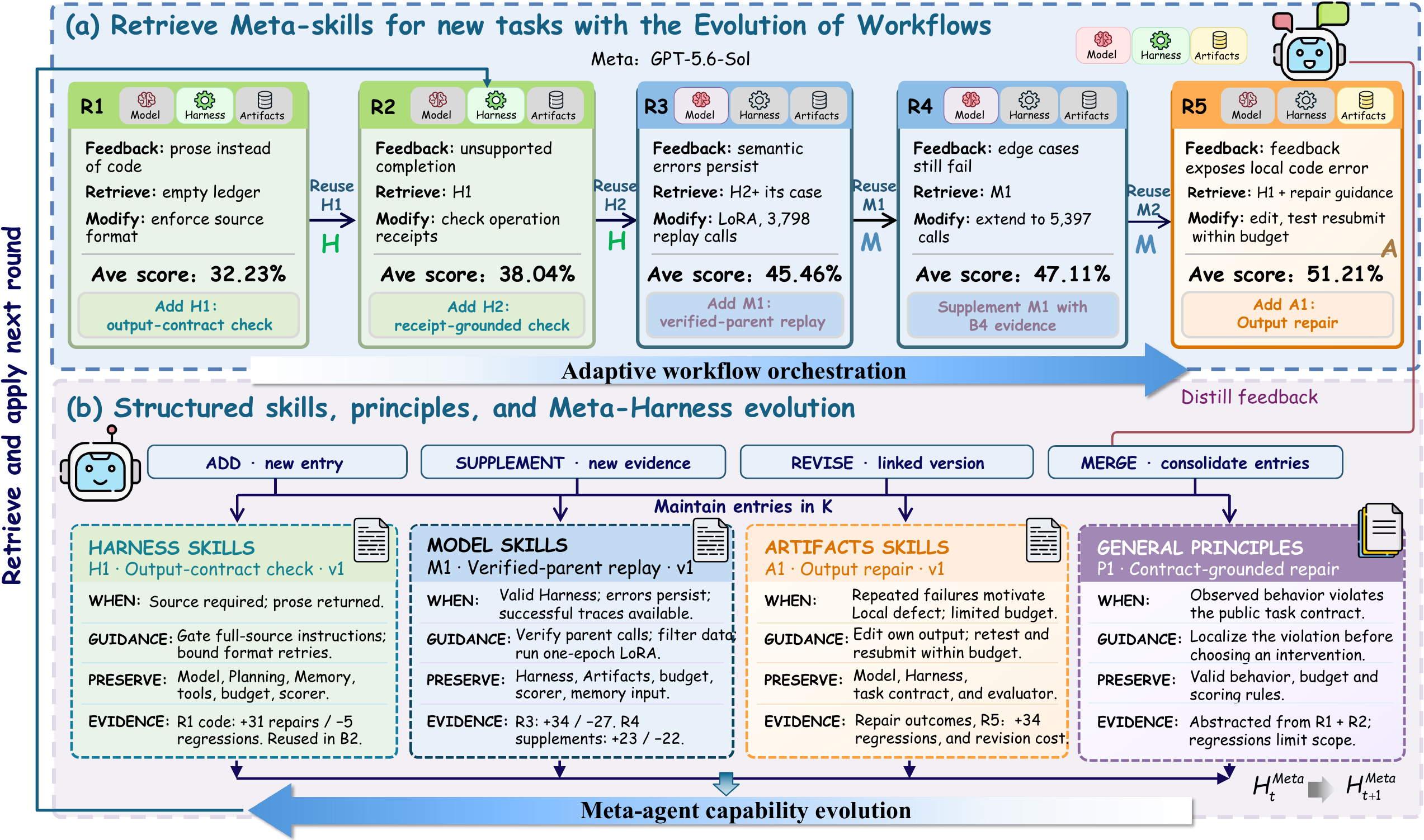}
    \caption{ Case study of five rounds of recursive self-improvement in RSIFlow. }
    \label{fig:case_study}
\end{figure*}

\subsection{Case Study}
\label{sec:case study}
Fig.~\ref{fig:case_study} illustrates five rounds
(R1--R5) of RSIFlow evolution, following a
Harness--Harness--Model--Model--Artifacts sequence.
In R1, the Meta-Agent addresses output-format
violations with an output-contract check (H1),
reaching 32.23\%. R2 builds on H1 to introduce
receipt-grounded verification (H2) for unsupported
completions, raising the score to 38.04\%.
To address persistent semantic errors, R3 shifts
to Model optimization with 3,798 replay calls
for LoRA fine-tuning, reaching 45.46\% and
distilling the model-specific skill M1.
R4 reuses and supplements M1 with new evidence,
expanding training to 5,397 calls and improving
the score to 47.11\%. In R5, execution feedback
and accumulated repair guidance support code revision
and resubmission within budget, reaching 51.21\%
while preserving the Model and Harness.
Across these rounds, skills and general principles
progressively refine the Meta-Agent's bottleneck
diagnosis and subsequent workflow orchestration.
This lineage demonstrates how adaptive workflow
orchestration and Meta-Harness evolution reinforce
each other, yielding an overall improvement of
18.98\%.

\section{Conclusion}

We presented \textbf{RSIFlow}, a recursive self-improvement framework coupling adaptive workflow orchestration with Meta-Agent evolution. Using accumulated execution feedback, the Meta-Agent identifies Task-Agent bottlenecks and selects Model, Harness, or Artifacts workflows. It distills successful and unsuccessful outcomes into reusable skills and principles to evolve its Meta-Harness. Across six benchmarks and two backbones, RSIFlow outperforms the strongest evaluated baseline in 11 of 12 settings. Ablations validate both mechanisms, while budget analyses show a favorable performance--budget trade-off. These results demonstrate that continual Meta-Agent evolution enhances system effectiveness alongside adaptive workflow orchestration.


\section*{Limitations}
RSIFlow is primarily evaluated using Qwen3-4B and Qwen3-8B as Task-Agent backbones. This setting is particularly relevant to resource-constrained agent deployments, where adaptive workflow orchestration enables targeted modifications to identified bottleneck components, while Meta-Agent evolution further improves bottleneck diagnosis and orchestration capabilities. However, whether these benefits remain significant with larger backbone models and at cluster scale remains unexplored. Future work will extend the evaluation to larger-scale settings to assess the scalability of RSIFlow.

RSIFlow couples the Task-Agent and Meta-Agent, evolving the Meta-Harness through reusable skills and general principles to improve workflow orchestration without updating the Meta-Agent's model parameters. However, this process introduces additional computational overhead, while adopting a lower-cost Meta-Agent may compromise evolution gains. Moreover, the stability of experience-driven Meta-Harness updates remains to be fully investigated. Future work will explore counterfactual analysis to identify Meta-Agent deficiencies, apply targeted local modifications, and investigate Meta-Agent parameter optimization to improve evolution efficiency and stability.

\bibliography{custom}

\clearpage

\appendix

\begingroup
\nolinenumbers
\raggedbottom
\setlength{\emergencystretch}{1em}

\newcommand{\rsiTableStyle}{%
  \small\renewcommand{\arraystretch}{1.15}%
  \setlength{\tabcolsep}{4pt}}

\newcounter{rsiAlgorithm}
\newenvironment{rsiAlgorithmBlock}[2]{%
  \par\addvspace{8pt}\noindent\begin{minipage}{\linewidth}
  \small\setlength{\parindent}{0pt}\setlength{\parskip}{1pt}%
  \hrule height .7pt\relax\nobreak\vspace{4pt}%
  \refstepcounter{rsiAlgorithm}\label{#2}%
  \noindent\textbf{Algorithm \thersiAlgorithm. #1}\par\nobreak
  \vspace{3pt}\hrule height .35pt\relax\nobreak\vspace{4pt}%
}{%
  \par\nobreak\vspace{3pt}\hrule height .7pt\relax
  \end{minipage}\par\addvspace{8pt}}
\newcommand{\rsiAlgLine}[2]{%
  \par\begingroup\raggedright
  \leftskip=#1em\relax\hangindent=1em\hangafter=1
  \noindent #2\par\endgroup}

\newcounter{rsiPrompt}
\newcommand{\rsiPromptImage}[3]{%
  \par\addvspace{8pt}\noindent
  \begin{minipage}{\linewidth}
  \centering\refstepcounter{rsiPrompt}\label{prompt:#1}\label{fig:prompt-#1}%
  \includegraphics[width=\linewidth]{figures/appendix_cards/prompt_#1.pdf}\par
  \vspace{3pt}{\footnotesize\raggedright\textbf{Prompt \thersiPrompt.} #3\par}%
  \end{minipage}\par\addvspace{8pt}}

\section{Experimental Setup}
\label{app:settings}
We evaluate self-improvement across code generation, tool use, and multi-hop question answering. This section describes the six benchmarks, the update mechanisms of the baselines, and the experimental configurations. Appendix~\ref{app:algorithms} gives the algorithms for the three intervention routes.

\subsection{Training and Reporting Data}
\label{app:data-distribution}
Our evolution pool contains 3,000 distinct training tasks,
partitioned into five task-disjoint batches of 600 tasks each.
Each round uses 200 code-generation tasks, 200 tool-use tasks,
and 200 multi-hop questions. The fixed reporting set contains
744 tasks. Table~\ref{tab:train-report-distribution} gives the
dataset-level allocation.

\paragraph{Code generation.}
\textbf{TACO} ~\cite{li2023tacotopicsalgorithmiccode} contains algorithmic programming problems that
require implementing a program from a natural-language
specification and satisfying its input/output contract.
We use 1,000 training problems from the TACO portion of the
\textbf{DeepCoder-Preview-Dataset} .
The selected tasks include standard-input/output and
function-based interfaces, with correctness assessed by
executing the submitted program against task-specific tests.
They provide execution trajectories and feedback for
improving program construction, interface handling, and
algorithmic reasoning. The allocation contributes 200 tasks
per evolution round; HumanEval+ and MBPP+ are reserved for
reporting evaluation.

\paragraph{Tool use.}
\textbf{EnvScaler} ~\cite{song2026envscalerscalingtoolinteractiveenvironments} provides executable, stateful environments
in which agents solve tasks through sequences of tool calls.
Each task specifies a user goal and an initial environment
state; the agent must choose tools, construct arguments, and
adapt its actions to returned observations. Task-specific
validation checks assess whether the resulting state satisfies
the goal.
We allocate 1,000 scenarios to training, comprising 824 from
the source dataset's \texttt{train} split and 176 from its
\texttt{heldout} split. Both source partitions are assigned to
the training role in our study; tool-use reporting uses
BFCL-v3 and ACEBench. Each evolution round contains 200
EnvScaler scenarios. Different training tasks may share an
environment, so the round-disjoint allocation refers to task
instances rather than environment families.

\paragraph{Multi-hop question answering.} \textbf{HotpotQA}~\citep{yang-etal-2018-hotpotqa} contains
questions that require combining evidence across Wikipedia
passages, including bridge and comparison questions.
\textbf{2WikiMultiHopQA}~\citep{ho-etal-2020-constructing}
requires following multi-hop relations across entities and
combining the retrieved facts into an answer. We select 500
questions from the official training split of each dataset.
The Task-Agent retrieves evidence through the search interface
and produces an answer, providing trajectories for improving
evidence retrieval and multi-step reasoning. Each round uses
100 HotpotQA and 100 2WikiMultiHopQA training questions.
Reporting uses a separate fixed set of 100 questions from
each dataset's official development split.

\paragraph{Allocation and data roles.}
Code, tool use, and multi-hop QA each account for one third of
the training pool, giving a domain ratio of $1{:}1{:}1$.
Within QA, HotpotQA and 2WikiMultiHopQA have equal weight.
The five training batches are fixed with split seed 42 and
contain distinct task IDs. These counts describe the task
allocation; candidate retries and reuse of eligible training
trajectories do not create additional distinct tasks.
Training and reporting use the separate source partitions
described above, with overlap checks and feedback isolation
specified in Appendix~\ref{app:evaluation-isolation}.

\begin{table*}[t]
\centering
\small
\renewcommand{\arraystretch}{1.12}
\setlength{\tabcolsep}{7pt}
\caption{Distribution of training and reporting tasks. Training
counts denote distinct tasks; the per-round column describes
each of the five evolution batches. A dash indicates that a
dataset is not used in that role.}
\label{tab:train-report-distribution}
\begin{tabularx}{\textwidth}{@{}l X r r r@{}}
\toprule
\textbf{Domain} & \textbf{Dataset}
& \shortstack{\textbf{Training}\\\textbf{total}}
& \shortstack{\textbf{Training}\\\textbf{per round}}
& \shortstack{\textbf{Reporting}\\\textbf{total}} \\
\midrule
Code & TACO (DeepCoder subset) & 1,000 & 200 & --- \\
     & HumanEval+             & --- & --- & 164 \\
     & MBPP+                  & --- & --- & 178 \\
\addlinespace[2pt]
     & \textit{Code subtotal} & \textit{1,000} & \textit{200} & \textit{342} \\
\midrule
Tool use & EnvScaler          & 1,000 & 200 & --- \\
         & BFCL-v3            & --- & --- & 152 \\
         & ACEBench           & --- & --- & 50 \\
\addlinespace[2pt]
         & \textit{Tool-use subtotal} & \textit{1,000} & \textit{200} & \textit{202} \\
\midrule
SearchQA & HotpotQA           & 500 & 100 & 100 \\
         & 2WikiMultiHopQA    & 500 & 100 & 100 \\
\addlinespace[2pt]
         & \textit{SearchQA subtotal} & \textit{1,000} & \textit{200} & \textit{200} \\
\midrule
\textbf{Total} & & \textbf{3,000} & \textbf{600} & \textbf{744} \\
\bottomrule
\end{tabularx}
\par\vspace{4pt}
\begin{minipage}{\textwidth}
\footnotesize\raggedright
Counts refer to task instances rather than generated
trajectories.
\end{minipage}
\end{table*}

\subsection{Benchmark Descriptions}
\label{app:metrics}
The reporting set contains 744 tasks: 342 code-generation tasks, 202 tool-use tasks, and 200 multi-hop questions. Each method is evaluated on the same task IDs with the same benchmark-specific scorer. Table~\ref{tab:app-benchmarks} summarizes the task counts and reported metrics.

\paragraph{Code generation.}
\textbf{HumanEval+} asks the agent to implement a Python function from its signature and natural-language specification. Its 164 tasks test functional correctness, including edge cases exposed by the augmented EvalPlus tests~\citep{NEURIPS2023_43e9d647}.
\textbf{MBPP+} asks for short Python programs from natural-language descriptions and illustrative tests, covering basic algorithms, data structures, and program construction. We use a frozen 178-task subset of MBPP+ v0.2.0.
For both benchmarks, a task succeeds when the final submitted program passes the base and augmented test suites. We report system pass@1: one final submission per task after the allowed agent execution. HumanEval+ uses v0.1.10, and both benchmarks use EvalPlus revision \texttt{26d6d00} with the recorded parser and process-isolation adapters.

\paragraph{Tool use.}
\textbf{BFCL-v3} evaluates function selection and argument construction in single-call, multiple-call, parallel-call, relevance, and multi-turn settings~\citep{pmlr-v267-patil25a}. These settings cover complementary aspects of tool use: selecting an appropriate function, supplying the required arguments, coordinating multiple calls, and determining whether an available tool is relevant to the request. The multi-turn tasks also test whether the agent maintains context across stateful tool interactions, connecting individual call decisions to the evolving interaction history. Our BFCL-v3 subset contains 152 tasks across 17 categories. We apply the native category-specific correctness checks and report category-macro accuracy, defined as the unweighted mean of the 17 category accuracies. Each category receives equal weight regardless of its task count.

\textbf{ACEBench} evaluates API use under Normal, Special, and Agent settings, including ordinary function calls, incomplete or ambiguous instructions, and multi-step interactions~\citep{chen-etal-2025-acebench}. Together, these settings examine how agents translate user requests into API actions under different instruction and interaction conditions. We use 50 English tasks, which produce 57 native scoring records. We report the native weighted score and separately retain complete-task success; these quantities use different denominators. The distinction matters because one task can contribute more than one native scoring record. Record-level scoring describes the evaluator's individual scoring units, whereas complete-task success summarizes outcomes at the original task level. We therefore specify both task and scoring-record counts when describing this evaluation.
\paragraph{Multi-hop question answering.}
\textbf{HotpotQA} requires combining evidence from multiple Wikipedia passages to answer bridge or comparison questions~\citep{yang-etal-2018-hotpotqa}. We use 100 development questions with the v1 fullwiki protocol.
\textbf{2WikiMultiHopQA} requires composing facts about multiple entities along multi-hop reasoning chains~\citep{ho-etal-2020-constructing}; we use 100 development questions and the original v1 evaluator.
These tasks test whether agents can connect dispersed evidence and resolve questions requiring relational reasoning.
For both benchmarks, the agent retrieves evidence through the search interface and submits an answer. The main metric is normalized answer-token F1, with exact match (EM) retained separately. The evaluator revisions are \texttt{3635853} for HotpotQA and \texttt{13800e5} for 2WikiMultiHopQA. 
Domain-level scores are arithmetic means of the two benchmark-level percentage scores in each domain. In particular, Tool Use averages BFCL-v3 category-macro accuracy and the ACEBench native weighted score. Overall is the arithmetic mean of the six benchmark-level percentage scores.

\begin{table}[t]
\centering\rsiTableStyle
\caption{Benchmarks and reporting metrics. ACEBench has 50 tasks and 57 native scoring records. QA exact match is retained separately.}
\label{tab:app-benchmarks}
\begin{tabularx}{\linewidth}{@{}l r >{\raggedright\arraybackslash}X@{}}
\toprule
\textbf{Benchmark} & \textbf{Tasks} & \textbf{Metric}\\
\midrule
HumanEval+ & 164 & System pass@1\\
MBPP+ & 178 & System pass@1\\
BFCL-v3 & 152 & Category-macro accuracy\\
ACEBench & 50 & Native weighted score\\
HotpotQA & 100 & Answer F1\\
2WikiMultiHopQA & 100 & Answer F1\\
\midrule
\textbf{Total} & \textbf{744} & ---\\
\bottomrule
\end{tabularx}
\end{table}

\subsection{Training--Evaluation Separation}
\label{app:evaluation-isolation}
We separate five rounds of training-stage evolution
from evaluation on the fixed 744-task reporting set.
For each evaluated evolution horizon $n$, we first
complete $n$ evolution rounds on training batches
and then evaluate the resulting frozen system.
When the Artifacts route is selected, the Task-Agent
revises task-specific outputs using execution
feedback, and the resulting outcomes provide
evidence for Meta-Skill evolution. Revised training
artifacts remain local to their original tasks and
are not transferred to unseen tasks. During
reporting, the selected Artifacts refinement
procedure can be applied to each evaluation task
through one additional feedback-guided revision.
The fixed-Artifacts ablation follows the same
five-round training protocol, with reporting-time
refinement performed only after training. 

The gain-based acceptance
rule in the evolution algorithm applies only to training-time
batches. During reporting, the model, Task-Harness, Meta-Harness, and skill library are frozen. When Artifacts refinement is invoked, the Meta-Agent uses the current task’s execution trace and permitted environment feedback to provide diagnostic guidance. The Task-Agent then revises its own output through the existing interface within the registered per-task budget. The final submission or execution outcome is scored using the benchmark-specific evaluator. Reporting scores, reference answers, and hidden-test outcomes are excluded from diagnostic guidance and output selection. During reporting, the skill library and Meta-Harness remain fixed across tasks. Permitted environment
feedback includes execution-interface signals such as
context-length limit errors and output-length or format
violations; it does not provide access to the reporting
evaluator for optimization or output selection. Reference
answers and hidden-test correctness signals are likewise
excluded from generation and output selection. All compared
methods receive the same environment feedback, tool permissions,
and per-task execution budgets.

We additionally checked the frozen training and reporting
manifests for overlap in task identifiers, recorded normalized
prompt hashes, and recorded task-group identifiers, finding no
matches between the two sets. This check establishes separation
under these matching criteria; it does not assess semantic
paraphrases or contamination in the backbone models'
pretraining data. Equal per-task execution budgets likewise do
not imply equal total evolution cost. Reporting-set scores and hidden correctness
signals are never used for workflow selection,
candidate acceptance, Meta-Skill maintenance,
Meta-Harness revision, or model selection.

\begin{table*}[t]
\centering\rsiTableStyle
\setlength{\tabcolsep}{4pt}
\caption{Baseline update configurations. M, H, and A denote Model, Harness, and Artifacts. Adaptive selection refers to choosing among Task-component intervention routes.}
\label{tab:app-shared}
\begin{tabularx}{\textwidth}{@{}l >{\centering\arraybackslash}X c c >{\centering\arraybackslash}X c c@{}}
\toprule
\textbf{Method} & \shortstack{\textbf{Task}\\\textbf{backbone}}
& \shortstack{\textbf{Model}\\\textbf{training}}
& \shortstack{\textbf{Harness}\\\textbf{updates}}
& \shortstack{\textbf{Editable Task}\\\textbf{components}}
& \shortstack{\textbf{Adaptive}\\\textbf{selection}}
& \shortstack{\textbf{Meta}\\\textbf{evolution}}\\
\midrule
SFT          & Qwen3-4B / 8B & Yes & No  & M       & No & No\\
GRPO         & Qwen3-4B / 8B & Yes & No  & M       & No & No\\
\addlinespace[2pt]
AFlow        & Qwen3-4B / 8B & No  & Yes & H       & No & No\\
MermaidFlow  & Qwen3-4B / 8B & No  & Yes & H       & No & No\\
\addlinespace[2pt]
HarnessForge & Qwen3-4B / 8B & Yes & Yes & M, H    & No & No\\
SIA          & Qwen3-4B / 8B & Yes & Yes & M, H    & Yes & No\\
HyperAgents  & Qwen3-4B / 8B & No  & Yes & Task code & No & Yes\\
\midrule
\rowcolor{gray!10}
\textbf{RSIFlow} & Qwen3-4B / 8B & Yes & Yes & \textbf{M, H, A} & \textbf{Yes} & \textbf{Yes}\\
\bottomrule
\end{tabularx}
\par\vspace{3pt}
\end{table*}

\subsection{Baseline Configurations}
\label{app:baselines}
We compare seven baselines using Qwen3-4B and Qwen3-8B~\citep{yang2025qwen3technicalreport}. Within each backbone scale, the comparison protocol uses a common initial checkpoint and Task-Harness, task IDs, and evaluators. Table~\ref{tab:app-shared} summarizes editable Task components, weight training, Harness updates, adaptive component selection, and Meta-Agent evolution. These dimensions distinguish how each baseline allocates adaptation across model parameters, task components, and orchestration.

\paragraph{Model updates.}
\textbf{SFT} trains on successful Task-Agent trajectories while keeping the Harness fixed.
\textbf{GRPO} samples groups of trajectories, scores their task outcomes, and updates the model using group-relative rewards~\citep{shao2024deepseekmathpushinglimitsmathematical}. Both methods improve the Task model without a component-selection mechanism.

\paragraph{Harness and joint updates.}
\textbf{AFlow} searches executable agent workflows through iterative proposal and evaluation~\citep{ICLR2025_5492ecbc}.
\textbf{MermaidFlow} represents workflows as graphs and searches their structure~\citep{zheng2025mermaidflowredefiningagenticworkflow}. These are the Harness-only configurations used in the main comparison.
\textbf{HarnessForge} combines feedback-guided Harness modification with model training in a prescribed co-adaptation procedure~\citep{chen2026harnessforgejointharnesspolicy}.
\textbf{SIA} uses execution feedback to choose between Harness and model updates~\citep{hebbar2026sia}.

\paragraph{Meta-Agent evolution.}
\textbf{HyperAgents} makes both the Task-Agent and Meta-Agent programs editable, allowing the improvement procedure itself to evolve~\citep{zhang2026hyperagents}. Its code-editing mechanism differs from parameter training.
\textbf{RSIFlow} selects among Model, Harness, and Artifacts interventions and updates its Meta-Harness through accumulated intervention skills. The comparison therefore separates access to multiple Task-update routes from the ability to revise the Meta-Agent that chooses and implements them.

\subsection{Hyperparameter Settings}
\label{app:implementation}
The main configuration uses Qwen3-4B or Qwen3-8B as the Task model and GPT-5.6 Sol as the Meta model. Each of the five evolution rounds permits at most two candidate attempts. Candidates share the same parent, training batch, and evaluator; the first valid candidate with positive paired gain is accepted. Meta model parameters remain fixed throughout evolution.

\begin{table}[ t]
\centering\rsiTableStyle
\caption{Main evolution settings. The separate 4B SFT/GRPO comparison appears in Appendix~\ref{app:rl-comparison}.}
\label{tab:app-rsi-config}
\begin{tabularx}{\linewidth}{@{}>{\raggedright\arraybackslash}p{0.43\linewidth}>{\raggedright\arraybackslash}X@{}}
\toprule
Parameter & Setting\\
\midrule
Task backbone & Qwen3-4B / Qwen3-8B\\
Meta backbone & GPT-5.6 Sol\\
Evolution rounds & 5\\
Attempts per round & At most 2\\
Default Model update & SFT\\
Candidate initialization & Frozen round parent\\
Acceptance & Valid and paired gain $>0$\\
\bottomrule
\end{tabularx}

\end{table}

Table~\ref{tab:app-training} lists the supplied 4B SFT/GRPO training settings. The separate comparison uses B1 for training and B2 for evaluation, as defined in Appendix~\ref{app:rl-comparison}; its evaluation temperature does not specify decoding for all main experiments. These settings provide context for interpreting the reported SFT and GRPO results, with B1 and B2 serving distinct roles in the training and evaluation stages of this specific experimental comparison.

\begin{figure*}[t]
\begin{rsiAlgorithmBlock}{RSIFlow Task--Meta evolution}{alg:rsi}
\rsiAlgLine{0}{\textbf{Input:} Task $T_0$, Meta $M_0$, skills $\mathcal K_0$, batches $\{\mathcal B_t\}$, rounds $R$, attempts $L$, and budget.}
\rsiAlgLine{0}{\textbf{Output:} Task $T_R$, Meta $M_R$, and skills $\mathcal K_R$.}
\rsiAlgLine{0}{\textbf{for} $t=0,\ldots,R-1$ \textbf{do}}
\rsiAlgLine{1}{Freeze parent $P=T_t$ and the evaluation protocol; execute $P$ on $\mathcal B_t$ to obtain $(\tau^-,f^-,J^-)$.}
\rsiAlgLine{1}{Initialize $T_{t+1}\gets P$ and experience $e_t\gets\varnothing$.}
\rsiAlgLine{1}{\textbf{for} $j=1,\ldots,L$ \textbf{do}}
\rsiAlgLine{2}{\textbf{if} budget exhausted \textbf{then break}.}
\rsiAlgLine{2}{Diagnose $(\tau^-,f^-,e_t)$, retrieve skills, and select $a\in\{\mathsf M,\mathsf H,\mathsf A\}$.}
\rsiAlgLine{2}{Record the proposed change, expected response, and preservation constraints; construct candidate $C$ from $P$.}
\rsiAlgLine{2}{\textbf{if} implementation checks fail: append failure to $e_t$ and \textbf{continue}.}
\rsiAlgLine{2}{Evaluate $C$ on $\mathcal B_t$; for Artifacts, score the Task-Agent's revised submissions. Record paired evidence in $e_t$.}
\rsiAlgLine{2}{\textbf{if} $C$ is valid and $J(C;\mathcal B_t)>J^-$: set $T_{t+1}\gets C$ and \textbf{break}.}
\rsiAlgLine{1}{\textbf{end for}}
\rsiAlgLine{1}{Distill component-specific skills and general principles from $e_t$, including failed attempts.}
\rsiAlgLine{1}{Maintain $\mathcal K_{t+1}$ and update the Meta-Harness to obtain $M_{t+1}$.}
\rsiAlgLine{0}{\textbf{end for}; \quad \textbf{return} $(T_R,M_R,\mathcal K_R)$.}
\end{rsiAlgorithmBlock}
\end{figure*}

\begin{table}[t]
\centering\rsiTableStyle
\caption{Recorded 4B training settings. $^\dagger$Evaluation temperature refers to the B1-training/B2-evaluation comparison in Table~\ref{tab:app-rl}. SFT lengths and batches are measured per eligible call; GRPO batches contain complete rollouts.}
\label{tab:app-training}
\begin{tabularx}{\linewidth}{@{}>{\raggedright\arraybackslash}X c c@{}}
\toprule
Parameter & 4B SFT & 4B GRPO\\
\midrule
Learning rate & $2\times10^{-6}$ & $2\times10^{-6}$\\
LoRA rank / alpha & 8 / 16 & 8 / 16\\
LoRA dropout & 0.05 & 0\\
Precision / ranks & BF16 / 8 & BF16 / 8\\
Maximum length & 16,384 & 32,768\\
Global batch & 8 calls & 8 trajectories\\
Epochs / passes & 1 & 1 per batch\\
Schedule & Linear & Constant\\
Warmup ratio & 0.03 & ---\\
Group size & --- & 2\\
Sampling temperature & --- & 1.0\\
Sampling top-$p$ & --- & 1.0\\
Clip $\epsilon$ / KL $\beta$ & --- & 0.2 / 0\\
Evaluation temperature$^\dagger$ & 0 & 0\\
\bottomrule
\end{tabularx}

\end{table}

\section{RSIFlow Implementation}
\label{app:algorithms}
\subsection{Task--Meta Evolution Loop}
\label{app:algorithm}

Algorithm~\ref{alg:rsi} gives the full loop; the main configuration uses $R=5$ and $L=2$.

The loop has two coupled outputs: an accepted Task state and an updated improvement mechanism. Every candidate starts from the same round parent, so rejected edits do not accumulate. Failed implementation attempts and non-improving evaluations still enter the experience record. On a new task batch, task-bound outputs are generated anew; reusable knowledge is carried by the Meta-Harness and accepted changes to the Task model or Harness. 
Algorithm~\ref{alg:rsi}  describes training-stage evolution
and candidate acceptance. Reporting-time artifact
refinement is executed separately using the
frozen system, as detailed in App. ~\ref{app:artifacts},
and does not contribute to subsequent
Meta-Agent updates.

\subsection{Workflow Orchestration}
\label{app:orchestration}
The Meta workflow connects six operations: evidence collection, diagnosis, skill retrieval, component intervention, paired evaluation, and skill integration. Diagnosis returns a bottleneck hypothesis with trajectory references; retrieval returns applicable procedures and counterexamples; intervention returns a candidate and its change record; evaluation returns paired outcomes; integration returns the next active skill set and the Action changes needed to use it.

A decision record binds the selected component to its expected behavioral effect and preservation constraints before construction. The Model branch receives eligible trajectories and a training configuration; the Harness branch receives a localized modification brief and interface contract; the Artifacts branch receives the Task's own output and permitted feedback. Evaluation determines Task acceptance, while the entire record informs Meta evolution. An accepted Task change and a revised Meta procedure are thus separate decisions within the same loop.

\subsection{Model Updates: SFT and GRPO}
\label{app:model}
The Model route updates the current model parameters while keeping the Task-Harness fixed and performing no direct Artifacts revision. Preserving Artifacts in this branch refers to the intervention scope; the candidate model generates its own outputs during evaluation. SFT reuses successful Task-Agent calls after provenance checks, deduplication, and length filtering. Its target is the final assistant response of each eligible call; environment observations provide context rather than supervised targets:
\begin{equation}
\mathcal L_{\mathrm{SFT}}
=-\frac{1}{|\mathcal D|}
 \sum_{(x,y)\in\mathcal D}\log p_\theta(y\mid x).
\end{equation}
The optimizer starts from the round parent. Successful calls from earlier training rounds can be reused when they satisfy the same eligibility rules, allowing prior successes to remain available for subsequent optimization steps.

GRPO collects groups of complete trajectories from a frozen behavior policy and assigns native task-success rewards $r_i\in\{0,1\}$. For a group of size $G$, the normalized advantage and token probability ratio are
\begin{equation}
\begin{aligned}
\widehat A_i&=\frac{r_i-\bar r}{\sigma_r+10^{-4}},\\
\rho_{i,j}(\theta)&=
\frac{p_\theta(y_{i,j}\mid h_{i,j})}
     {p_{\mathrm{old}}(y_{i,j}\mid h_{i,j})}.
\end{aligned}
\end{equation}
Here, $h_{i,j}$ includes the task and interaction history available at that token. With $L_i$ denoting the number of response tokens across all calls in trajectory $i$, the recorded 4B implementation minimizes
\begin{equation}
\begin{aligned}
\mathcal L_{\mathrm{GRPO}}
={}&-\frac{1}{G}\sum_{i=1}^{G}\frac{1}{L_i}
       \sum_{j=1}^{L_i}
       \min\bigl\{\rho_{i,j}\widehat A_i,\\
&\qquad
\operatorname{clip}(\rho_{i,j},1-\epsilon,1+\epsilon)
\widehat A_i\bigr\}.
\end{aligned}
\label{eq:app-grpo}
\end{equation}
This configuration sets the KL coefficient to zero and excludes observation tokens from the loss. Zero-variance reward groups remain in rollout accounting but contribute no policy-gradient update. Chosen-token log probabilities come from the frozen sampling policy.

\begin{rsiAlgorithmBlock}{Model update}{alg:model}
\rsiAlgLine{0}{\textbf{Input:} Parent $(\theta,H,A)$, eligible training data, objective, and budget.}
\rsiAlgLine{0}{1. Freeze the parent and register the training configuration.}
\rsiAlgLine{0}{2. For SFT, select successful calls, deduplicate, and construct assistant-token targets.}
\rsiAlgLine{0}{3. For GRPO, sample trajectory groups and compute rewards, advantages, and token masks.}
\rsiAlgLine{0}{4. Optimize the selected loss starting from $\theta$.}
\rsiAlgLine{0}{5. Verify the checkpoint, serving binding, and unchanged $H,A$.}
\rsiAlgLine{0}{\textbf{Return:} $(\theta',H,A)$ for paired evaluation, or failure if verification fails.}
\end{rsiAlgorithmBlock}

\subsection{Harness Updates}
\label{app:harness}
Harness intervention changes how the agent executes tasks. Meta localizes a recurring execution defect, refines a modification brief, generates a complete implementation, and validates it. The functional components are Action, Skill, and Planning; concrete modules may implement context organization, action handling, memory retrieval, or verification. Localization identifies the mechanism to change, rather than merely listing files. Executable checks establish that a candidate can run. Acceptance additionally requires a positive paired task gain. An output-contract check can ensure that a coding task receives source code, whereas a receipt-grounded completion check can ensure that a tool operation has actually succeeded. These are changes to reusable execution behavior.
\begin{rsiAlgorithmBlock}{Harness update}{alg:harness}
\rsiAlgLine{0}{\textbf{Input:} Parent $(\theta,H,A)$, execution evidence, skills, and repair allowance.}
\rsiAlgLine{0}{1. Localize the execution defect and successful behavior to preserve.}
\rsiAlgLine{0}{2. Select components $\mathcal I$ and specify the change, expected response, and constraints.}
\rsiAlgLine{0}{3. Generate $\widehat H^{\mathcal I}$ and set $H'=H[\mathcal I\leftarrow\widehat H^{\mathcal I}]$.}
\rsiAlgLine{0}{4. Check syntax, imports, construction, interfaces, and targeted behavior. Repair and recheck within allowance.}
\rsiAlgLine{0}{5. Verify unchanged $\theta,A$ and unselected Harness components.}
\rsiAlgLine{0}{\textbf{Return:} $(\theta,H',A)$ for paired evaluation, or failure if checks fail.}
\end{rsiAlgorithmBlock}











\subsection{Artifacts Updates}
\label{app:artifacts}
An output-artifact intervention changes the Task's deliverable while keeping its model and Harness fixed. Meta identifies the defect and supplies diagnostic guidance; the Task-Agent revises its own output through the existing interface. For a coding task, the inputs are the specification, submitted program, permitted execution feedback, and remaining budget. The output is a revised program, not a replacement answer authored by Meta.

\paragraph{Training and Evaluation Protocol.}
Artifact refinement serves different roles during
training and evaluation. During training, the Meta-Agent
guides the Task-Agent to revise task-specific outputs
using execution feedback, and the resulting changes
provide evidence for subsequent Meta-Skill maintenance.
These revised outputs remain local to their original
tasks and are not transferred to unseen tasks. During
evaluation, when the Artifacts route is selected, the
Task-Agent generates an initial output and receives
one additional opportunity to revise it using permitted
environment feedback and the Meta-Agent's previously
accumulated guidance. The Model, Task-Harness,
Meta-Harness, and Meta-Skill library remain fixed
throughout this process. The benchmark evaluator
scores only the final submission, and its scores
or hidden-test results are never exposed to the
revision procedure. Consequently, the resulting
performance gain reflects feedback-guided correction
of evaluation outputs rather than the generalization
of repaired training artifacts.

\begin{rsiAlgorithmBlock}{Task-output refinement}{alg:artifact_refinement}
\rsiAlgLine{0}{\textbf{Input:} Task $x$, initial output $o^{(0)}$, execution trace $\tau$, permitted feedback $f$, fixed Task $(\theta,H)$, Meta-Agent $M$, skill library $K$, mode $m\in\{\mathrm{train},\mathrm{report}\}$, and revision allowance $L_m$.}
\rsiAlgLine{0}{1. Bind $(o^{(0)},\tau,f)$ to task $x$ and initialize $o^*\leftarrow o^{(0)}$.}
\rsiAlgLine{0}{2. Meta retrieves applicable skills from $K$, localizes the defect, and produces diagnostic guidance $\delta$.}
\rsiAlgLine{0}{3. The Task-Agent generates candidate revisions $o^{(j)}$ using $(x,o^{(0)},\tau,f,\delta)$ while keeping $\theta,H$ unchanged. Set $L_{\mathrm{report}}=1$.}
\rsiAlgLine{0}{4. Check output validity, permitted local tests, and execution constraints. Retain a valid revision as $o^*$; otherwise preserve the original output.}
\rsiAlgLine{0}{5. During training, return the revised output and evidence for paired training-batch evaluation and subsequent Meta-Skill maintenance.}
\rsiAlgLine{0}{6. During reporting, submit the final output for benchmark scoring while keeping $\theta,H,M,K$ frozen and excluding evaluator results from further updates.}
\rsiAlgLine{0}{\textbf{Return:} Final output $o^*$ and task-local revision evidence.}
\end{rsiAlgorithmBlock}

For example, a permitted test may expose an omitted empty-input branch. The Task-Agent inspects the failing path, adds the required branch, and resubmits. This illustrates the interface rather than an additional measured benchmark case. Hidden evaluator answers are never revision inputs. Stateful tool tasks require execution or replay against the registered environment state: editing a final textual claim cannot establish that an operation succeeded.
\begin{table*}[t]
\centering\rsiTableStyle
\caption{Five-round HHMMA case: Task interventions, accumulated skills, and reported average scores. Skill entries summarize the reported procedures; A1 denotes feedback-guided output revision at R5.}
\label{tab:app-hhmma-changes}
\label{tab:app-hammh-changes}
\label{tab:app-skills}
\begin{tabularx}{\textwidth}{@{}l c r >{\raggedright\arraybackslash}X >{\raggedright\arraybackslash}X@{}}
\toprule
\textbf{Round} & \textbf{Route} & \shortstack{\textbf{Average}\\\textbf{(\%)}} & \textbf{Task intervention} & \textbf{Skill update and reuse}\\
\midrule
R1 & H & 32.23 & Add output-contract verification. & H1: distinguish interface errors from semantic failures.\\
\addlinespace[3pt]
R2 & H & 38.04 & Add receipt-grounded checks for unsupported completions. & H2: reuse H1's contract-checking criterion for execution evidence.\\
\addlinespace[3pt]
R3 & M & 45.46 & Apply LoRA replay with 3,798 eligible calls. & M1: consolidate recurring behavior through successful-call replay.\\
\addlinespace[3pt]
R4 & M & 47.11 & Extend replay to 5,397 eligible calls. & Reuse M1 and supplement it with new evidence.\\
\addlinespace[3pt]
R5 & A & 51.21 & Meta-guided revision and resubmission of the Task-Agent’s own output using execution feedback; Model and Harness remain fixed. & A1: output repair using prior guidance and current feedback.\\
\bottomrule
\end{tabularx}
\end{table*}
\paragraph{Persistence and cross-task reuse.}
The Artifacts component in the main-text Task state denotes task-bound deliverables. Writing it as $A_{t,x}$ makes the task index explicit: a revised output is retained for continued work on task $x$, while a new task $x'$ produces its own $A_{t,x'}$. Cross-task reuse occurs through the Meta-Skill library $\mathcal K_t$, which retains repair conditions, procedures, outcomes, and limitations. Meta retrieves this guidance to select and guide revision on a later task; the Task-Agent generates and revises that task's own output. The R5 case is summarized in Appendix~\ref{app:HHMMA}.

A Task-Harness skill implements reusable task-execution behavior, whereas a Meta-Skill guides diagnosis and intervention. Updating the former is a Harness intervention; distilling the latter is a subsequent Meta-level update. Persistent resource files or memory are therefore distinct from the task-bound output revised by the Artifacts route. Table~\ref{tab:app-component-scope} summarizes these boundaries.

\begin{table*}[t]
\centering\rsiTableStyle
\caption{Task-component intervention scope and persistence. Meta-Skill maintenance follows the intervention and is recorded separately. Task contracts, evaluators, and registered execution limits are preserved across all routes.}
\label{tab:app-component-scope}
\begin{tabularx}{\textwidth}{@{}l *{4}{>{\raggedright\arraybackslash}X}@{}}
\toprule
\textbf{Component} & \textbf{Editable object} & \textbf{Retained across rounds} & \textbf{Use on a new task} & \textbf{Preserved scope}\\
\midrule
Model & Task-model parameters or adapters. & Accepted checkpoint or adapter. & Generate fresh outputs with the updated model. & Harness unchanged; no direct output revision.\\
\addlinespace[3pt]
Harness & Action, Skill, or Planning modules. & Accepted execution logic and configuration. & Execute the new task using the revised Harness. & Model unchanged; no direct output revision.\\
\addlinespace[3pt]
Artifacts & Current code, structured submission, or answer. & Revised output for the same task; reusable experience enters Meta-Skills separately. & Generate a fresh output and reuse applicable revision guidance. & Model parameters and Harness unchanged.\\
\bottomrule
\end{tabularx}
\end{table*}

\paragraph{Examples across task domains.}
\textbf{Code:} a permitted test reveals an omitted boundary case; Task-Agent edits the program, retests, and resubmits the source.
\textbf{Tool Use:} an API reports a missing argument; the Task-Agent corrects and resubmits its structured call through the existing interface, with success established by the resulting environment receipt.
\textbf{SearchQA:} retrieved evidence reveals an entity mismatch; the Task-Agent revises its answer using that evidence while keeping the retrieval Harness fixed.
These examples illustrate the intervention scope, rather than additional measured cases.

\section{Additional Results and Case Studies}
\label{app:analysis}

We report the Model-update comparison, additional rollout-budget results, and two complementary views of Task--Meta evolution.

\subsection{SFT and GRPO Comparison}
\paragraph{Training configuration.}
Table~\ref{tab:app-training} gives the recorded 4B configurations. The two replay updates described in the main HHMMA case use 3,798 and 5,397 eligible calls. The separate SFT/GRPO comparison uses 2,780 eligible SFT calls from 1,200 source tasks, versus 6,000 GRPO trajectories from 3,000 tasks. GRPO carries the adapter and optimizer across five 600-task batches. The supplied optimization trace contains 100 optimizer updates; the trajectory count describes collection, while the trace describes recorded optimization steps.
\begin{figure}[t]
\centering
\includegraphics[width=\linewidth]
{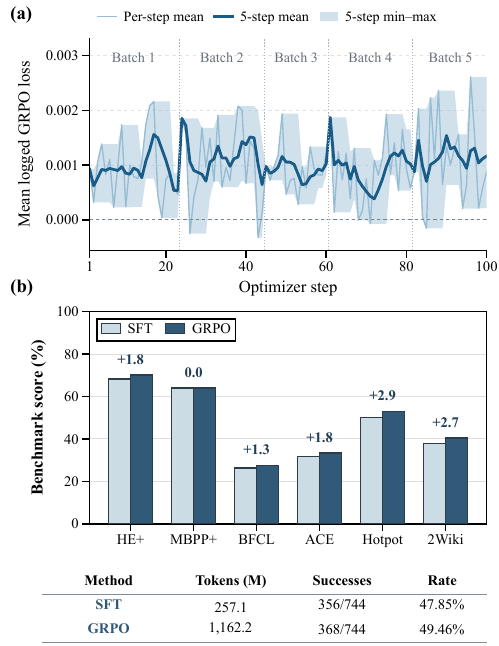}
\caption{Model-update comparison.
(a) GRPO loss with a five-step mean and min--max band.
(b) Benchmark scores, rollout tokens, and success on
744 tasks. Labels show GRPO--SFT gains (\%). }
\label{fig:app-grpo-sft}
\end{figure}

\begin{figure*}[t]
\centering
\includegraphics[width=\textwidth]
{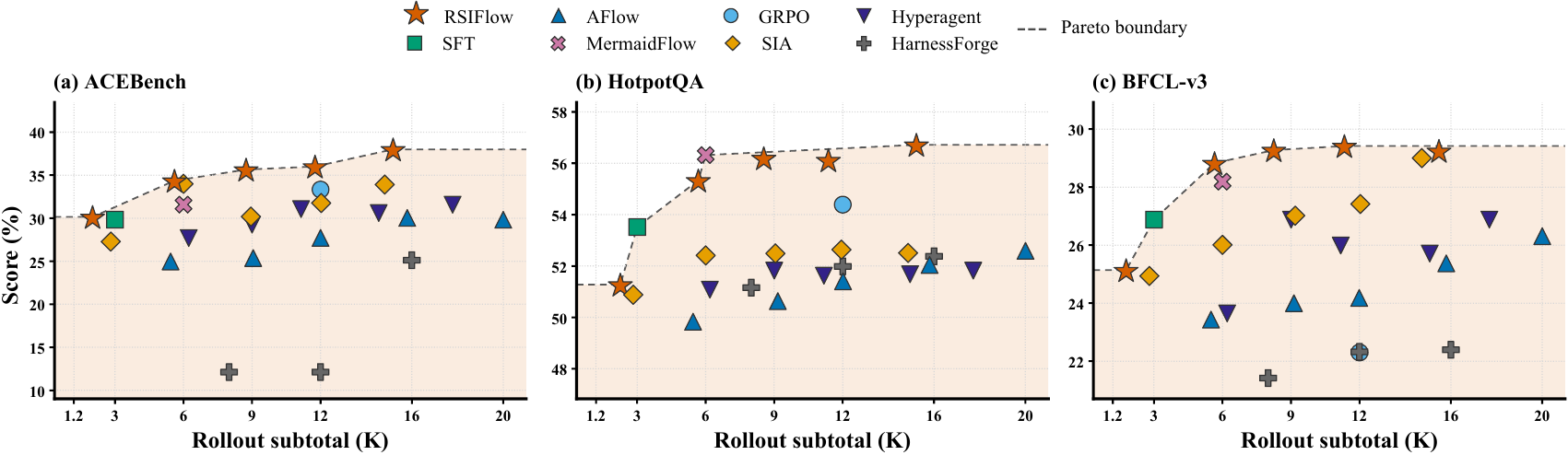}
\caption{Rollout-budget--performance trade-offs on
ACEBench, HotpotQA, and BFCL-v3, complementing
the main-text budget analysis. Stars denote RSIFlow; dashed lines trace
the empirical Pareto boundary. K denotes 1,000 rollouts.}
\label{fig:app-budget}
\end{figure*}

\begin{table}[t]
\centering\rsiTableStyle
\caption{Measured 4B B1-training/B2-evaluation results. Scores are percentages; complete-task success uses QA EM and complete ACE task success. }
\label{tab:app-rl}
\begin{tabularx}{\linewidth}{@{}>{\raggedright\arraybackslash}X r r@{}}
\toprule
Metric & SFT-1200 & GRPO-3000\\
\midrule
HumanEval+ pass@1 & 68.29 & 70.12\\
MBPP+ pass@1 & 64.04 & 64.04\\
BFCL-v3 category-macro accuracy & 26.31 & 27.61 \\
ACEBench weighted score & 31.57 & 33.33\\
HotpotQA answer F1 & 50.06 & 52.91\\
2Wiki answer F1 & 37.79 & 40.45\\
\midrule
Complete-task successes & 356/744 & 368/744\\
Complete-task success rate & 47.85 & 49.46\\
\bottomrule
\end{tabularx}
\end{table}

\paragraph{Comparison protocol.}
\label{app:rl-comparison}
Both methods start from the original Qwen3-4B checkpoint, train with the same B1 Harness, and report with the same B2 Harness. B1 denotes the complete-source/shape-check Harness, and B2 its receipt-check successor. These labels identify training and evaluation configurations in this separate SFT/GRPO comparison; its SFT-1200 and GRPO-3000 results are not the SFT and GRPO baseline rows in the main results table. SFT selects 2,780 calls from 300 successful tasks among 1,200 source tasks; GRPO samples two trajectories for each of 3,000 tasks. Both complete all 744 reporting tasks. GRPO reaches 368 complete-task successes versus 356 for SFT, a difference of 12 tasks or 1.61\%. This complete-task success rate uses binary outcomes across 744 tasks and differs from the benchmark-score average reported as Overall in the main results. This comparison varies the training objective and data budget jointly.

\paragraph{Training dynamics and rollout cost.}
Figure~\ref{fig:app-grpo-sft} complements the final
results with training dynamics and data-collection cost.
Panel~(a) summarizes 100 optimizer steps across five
rollout batches; the five-step statistics are computed
within each batch and describe local loss fluctuations.
Panel~(b) shows higher GRPO scores on five benchmarks
and an unchanged MBPP+ score. Complete-task success increases
from 47.85\% to 49.46\%, while recorded rollout usage
increases from 257.1M to at least 1,162.2M tokens.
These totals count input and output tokens used to
collect training trajectories, including the historical
collection cost attributed to SFT; evaluation, Meta
calls, and optimizer computation are excluded.
SFT reuses trajectories from 1,200 source tasks without
new sampling, whereas GRPO collects two trajectories
per task from 3,000 tasks. Complete-task success increases by 1.61\%, accompanied by substantially greater trajectory collection costs. The two configurations
therefore illustrate different performance--collection
cost trade-offs within the Model intervention route.

\subsection{Rollout-Budget Analysis}
\label{app:budget-analysis}

Figure~\ref{fig:app-budget} provides an additional view of
the rollout-budget trade-offs on ACEBench, HotpotQA, and BFCL-v3. The horizontal axis measures rollout count, and the vertical axis reports each benchmark's score. The empirical Pareto boundary identifies configurations
for which no evaluated alternative achieves an equal
or higher score with an equal or smaller rollout budget
and a strict improvement in at least one dimension. RSIFlow attains the highest observed score on all three
benchmarks, with several configurations lying on the
empirical Pareto boundary. The budget response differs
across tasks: BFCL-v3 scores level off at larger budgets,
ACEBench improves further toward the largest RSIFlow
budget, and HotpotQA exhibits intermediate fluctuations
before reaching its highest score. Together with the
main-text results, these observations show that the
benefit of additional rollouts is benchmark-dependent
and motivate allocating improvement effort according
to the current task response.

\subsection{Five Rounds of Adaptive Intervention}
\label{app:HHMMA}
\label{app:HAMMH}
\label{app:task-evolution}
\label{app:round-skills}
The main-text case follows Harness--Harness--Model--Model--Artifacts (HHMMA). Table~\ref{tab:app-hhmma-changes} links each reported average to the intervention and the knowledge available for the next round. Figure~\ref{fig:app-hhmma} presents the same sequence. Meta$_0$ acts in R1 and Meta$_4$ in R5; each intervention's feedback updates the Meta version used next.

\paragraph{R1--R2: repair execution contracts through Harness updates.}
R1 addresses output-format violations by introducing the output-contract check H1, achieving an average score of 32.23\%. R2 reuses H1 and prior experience to address unsupported completions through receipt-grounded verification H2, increasing the score to 38.04\%. The two rounds apply a shared contract-checking criterion to different execution defects: submitting the required deliverable and verifying that a claimed operation succeeded.

\paragraph{R3--R4: consolidate recurring behavior in the model.}
As semantic errors persist, R3 shifts to Model optimization. R3 introduces M1 through 3,798 replay calls; R4 reuses that procedure with 5,397 calls. The second application supplements M1 rather than creating a duplicate skill. The reported scores are 45.46 and 47.11.

\paragraph{R5: repair the current output within budget.}
Finally, R5 shifts to Artifacts refinement. The Task-Agent uses execution feedback and prior repair guidance to edit, test, and resubmit its current code within budget, keeping the Model and Harness fixed. This targets a local output defect without an additional model-training or Harness-reconstruction step. The final reported average is 51.21\%, 18.98\% above R1. This measures the span from the first reported intervention to the fifth. Reusable repair guidance enters Meta-Skills; the revised code remains specific to the current task. During the fifth training round, the Artifacts
route refines training-task outputs and uses
the resulting revision evidence to develop
the artifact-specific Meta-Skill A1. During
reporting, the frozen system applies this
refinement guidance to newly generated
evaluation-task outputs, allowing at most
one additional revision based on permitted
task-local feedback before final scoring.
Thus, the improvement from R4 to R5
reflects the effectiveness of feedback-guided
inference-time correction supported by
accumulated Meta-Agent experience, rather
than the direct transfer of revised training
artifacts.

\begin{figure}[t]
\centering
\includegraphics[width=\linewidth]{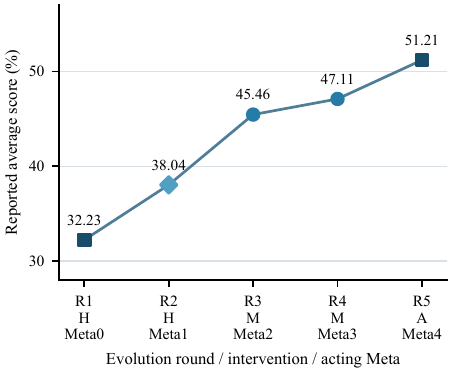}
\caption{HHMMA evolution across five rounds. H, A, and M denote the selected intervention; points show the reported average scores.}
\label{fig:app-hhmma}
\label{fig:app-hammh}
\end{figure}

\subsection{Selected Meta Versions on a Common Task Initialization}
\label{app:meta-compatibility}
The main text also reports a different comparison: Meta versions modify a common initial Task-Agent rather than successively inheriting the Task states of the HHMMA case. Table~\ref{tab:app-meta-domain} reproduces the reported domain scores. Its intervention labels are H--H--M--A--H because they belong to this separate comparison, not to the five-round HHMMA trajectory.

\begin{table}[ t]
\centering\rsiTableStyle
\caption{Selected Meta versions acting on a common initial Task-Agent. Each score is a macro-average over the two benchmarks in that domain. Scores are independent of the sequential HHMMA lineage.}
\label{tab:app-meta-domain}
\begin{tabularx}{\linewidth}{@{}l c *{3}{>{\centering\arraybackslash}X}@{}}
\toprule
Meta & Route & Code & Tool & QA\\
\midrule
Initial Task & --- & 48.25 & 29.70 & 34.52\\
Meta0 & H & 48.30 & 30.40 & 44.62\\
Meta1 & H & 46.32 & 31.16 & 45.04\\
Meta2 & M & 67.90 & 31.51 & 43.99\\
Meta3 & A & 67.08 & 31.83 & 50.58\\
Meta4 & H & 68.86 & 30.84 & 54.13\\
\bottomrule
\end{tabularx}

\end{table}

Meta0 and Meta4 both select Harness changes, yet their reported Code scores differ by 20.56\% and their QA scores by 9.51\%. Selecting the same component therefore does not determine the quality of the proposed modification. Meta3 has the highest reported Tool Use score in this table, showing that the ordering of Meta versions is domain-dependent. Across the selected versions, Code and QA show larger score variation than Tool Use, indicating that the observed performance differences between Meta versions are more pronounced in some domains than others. The sequential case tracks evolving Task states; the common-initialization comparison evaluates different Meta versions on the same Task parent.

\subsection{Meta-Agent Backbone Sensitivity}
\label{app:meta-backbone-sensitivity}


Figure~\ref{fig:meta-backbone-cumulative} complements the main-text sensitivity
analysis with a cumulative-token view. From
Round 1 to Round 5, HumanEval+ pass rates increase
from 42.07\% to 53.66\% for Claude-Sonnet-5,
from 44.51\% to 49.39\% for DeepSeek-V4.1-Flash,
and from 43.90\% to 68.29\% for GPT-5.6-Sol,
corresponding to gains of 11.59\%, 4.88\%,
and 24.39\%. Their final cumulative Meta-Agent
token counts are 177.7M, 211.2M, and 252.6M,
respectively. GPT-5.6-Sol reaches 60.37\% after
four rounds using 91.8M tokens; its fifth round
consumes an additional 160.8M tokens for a further
7.93\% improvement. Claude-Sonnet-5 achieves a
higher final pass rate than DeepSeek-V4.1-Flash
with fewer cumulative tokens, while
DeepSeek-V4.1-Flash exhibits a small third-round
decline. These trajectories illustrate how the
observed performance--token trade-off varies
across Meta-Agent backbones and evolution rounds.

\begin{figure*}[t]
    \centering
    \includegraphics[width=\textwidth]{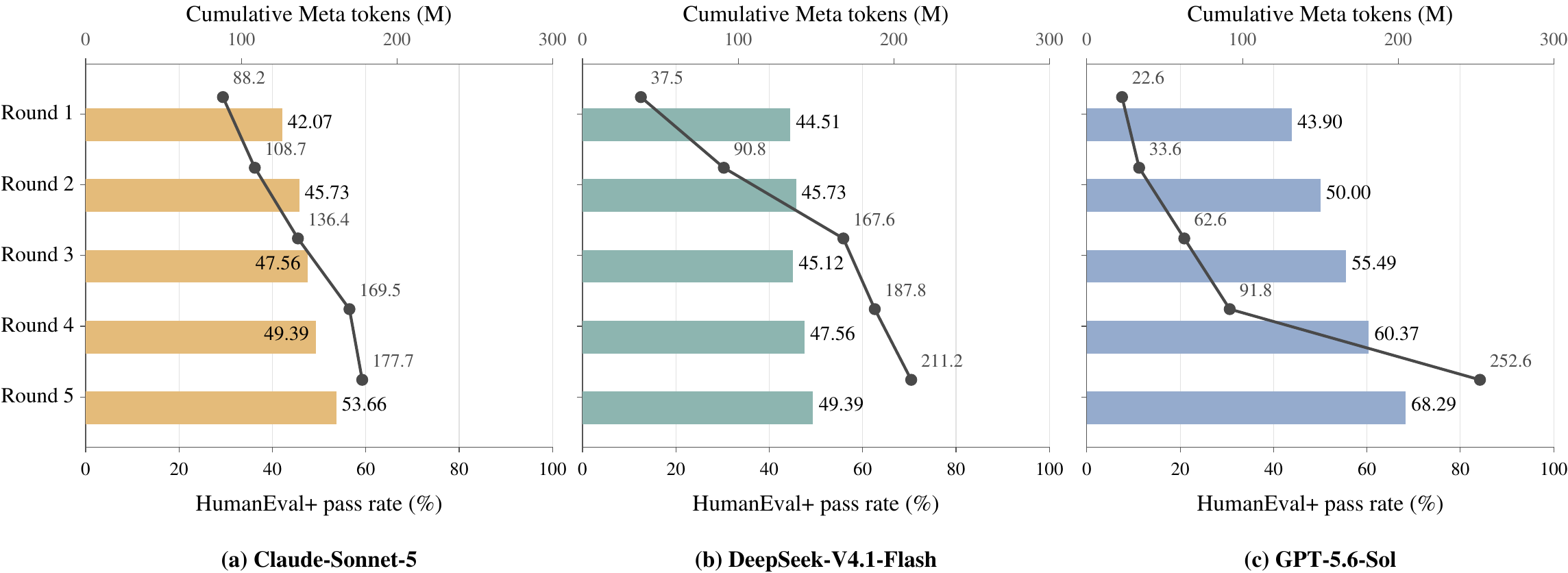}
    \caption{Meta-Agent backbone sensitivity on HumanEval+ over five evolution rounds. Bars show pass rates using the bottom axes, while lines show cumulative Meta-Agent token consumption using the top axes. Cumulative counts sum token usage through the corresponding round; M denotes one million tokens.}
    \label{fig:meta-backbone-cumulative}
\end{figure*}
\section{Meta-Harness Evolution}
\label{app:skills}
Interventions produce both Task candidates and experience for improving the Meta-Agent. This section describes skill representation, maintenance, and integration, detailing how intervention experience is organized and incorporated into the ongoing improvement of the Meta-Agent. Prompt excerpts are collected in Appendix~\ref{app:prompts}.

\subsection{Failure Diagnosis and Intervention}
\label{app:skill-mapping}
A visible error does not uniquely determine an intervention. We represent the diagnosed state as
\begin{equation}
z_t=(c_t,\epsilon_t,\ell_t,s_t,\Gamma_t),
\end{equation}
where $c_t$ is the task contract, $\epsilon_t$ the observed error, $\ell_t$ its localized execution stage, $s_t$ the current Task state, and $\Gamma_t$ the behaviors to preserve. Invalid code, for example, can arise from a missing submission check, a semantic weakness, or a locally repairable program defect.

The library contains general principles $\mathcal P_t$ and component-specific procedures $\mathcal S_t^a$:
\begin{equation}
\mathcal K_t=\mathcal P_t\cup
\bigcup_{a\in\{\mathsf M,\mathsf H,\mathsf A\}}\mathcal S_t^a.
\end{equation}
A skill is represented by
\begin{equation}
k=(\Omega_k,a_k,\rho_k,\Gamma_k,\widehat r_k,E_k),
\label{eq:app-skill-tuple}
\end{equation}
with applicability predicate $\Omega_k$, intervention scope $a_k$, procedure $\rho_k$, preservation requirements $\Gamma_k$, expected response $\widehat r_k$, and supporting or contradicting evidence $E_k$. Eligible guidance is
\begin{equation}
\mathcal K_t(z)=\{k\in\mathcal K_t:\Omega_k(z)=1\}.
\end{equation}
Retrieval proposes matches; Meta checks their applicability against the current traces using the stored conditions and exclusions.

With $K=\mathcal K_t(z)$, the decision factorizes into route selection and modification construction:
\begin{equation}
\begin{aligned}
\pi_t(a,\delta\mid z,K)
={}&\pi_t(a\mid z,K)\\
&\cdot\pi_t(\delta\mid a,z,K).
\end{aligned}
\label{eq:app-factorization}
\end{equation}
General principles guide diagnosis and comparisons across intervention routes, while component-specific procedures guide feasible edits within a route. This factorization describes their roles in a conditional decision process; prompting and retrieval implement the process without a separately fitted routing classifier.

\paragraph{An explicit retrieval implementation.}
The case study skill retriever provides a concrete implementation of retrieval. It scores lexical matches between query tokens $Q(z)$ and searchable skill tokens $B(k)$, emphasizing error tokens $F(z)$:
\begin{equation}
\begin{aligned}
r(k,z)={}&|Q(z)\cap B(k)|\\
&+3|Q(z)\cap B(k)\cap F(z)|.
\end{aligned}
\end{equation}
Each matched error token contributes four units to the score, whereas other matched query tokens contribute one unit each. This weighting favors records sharing error terms with the query while preserving general lexical overlap as a complementary signal for skill retrieval.
Domain aliases are normalized, record IDs break ties, and the returned context is bounded to 24,000 characters with at most one record per bucket. Full records remain accessible by ID. This score retrieves evidence; it does not estimate an intervention's numerical payoff.

\subsection{Paired Evidence, Skill Maintenance, and Abstraction}
\label{app:skill-maintenance}
For an intervention evaluated on the same task set, a local evidence unit is
\begin{equation}
u_i=(z_i,a,\delta,\tau_i^-,\tau_i^+,y_i^-,y_i^+).
\end{equation}
Trajectory fragments align at shared subgoals, tool events, or error signatures, so alignment does not require equal trace lengths. Outcomes retain full-task success indicators $y_i^\pm$. Define
\begin{equation}
n_{uv}=\sum_i\mathbb{I}[y_i^-=u,\ y_i^+=v].
\end{equation}
For the special case of unweighted binary success on $N$ tasks, the paired gain is
\begin{equation}
\Delta J=\frac{n_{01}-n_{10}}{N}.
\label{eq:app-paired-decomposition}
\end{equation}
This decomposition describes binary success. More generally, candidate gain is
\begin{equation}
\Delta J=J(T';\mathcal B_t)-J(T_t;\mathcal B_t).
\end{equation}
F1 and native weighted scores retain their corresponding scoring rules; domain-macro scores are computed within each domain before averaging. Retained successes identify behavior to preserve, while retained failures identify unresolved mechanisms. Local fragments explain the response but do not replace full-batch scoring.

\begin{table}[t]
\centering\rsiTableStyle
\caption{Maintenance operations for component skills and general principles. Original cases remain traceable after active content changes.}
\label{tab:app-maintenance}
\begin{tabularx}{\linewidth}{@{}l >{\raggedright\arraybackslash}X@{}}
\toprule
Operation & Evidence-driven change\\
\midrule
Add & Create an entry for an uncovered strategy, with conditions and source cases.\\
Supplement & Attach support or counterexamples to an existing strategy.\\
Revise & Update a condition, step, or invariant in a linked successor version.\\
Merge & Consolidate duplicate or superseded entries in the active view.\\
\bottomrule
\end{tabularx}

\end{table}

A general principle abstracts a shared decision criterion from component-specific cases. For example, completion requiring task-appropriate evidence connects a source-code contract check with a tool-receipt check, while their implementations remain separate. The abstraction records source cases and applicability conditions; reuse contributes another response record rather than automatically expanding its scope.

\subsection{Component-Specific Skills and a General Principle}
\label{app:skill-card-examples}

The cards in Appendix~\ref{app:prompts} present three component-specific procedures
and one general selection principle. The Harness and Model cards
summarize stored procedures and their observed feedback.
The Artifacts card formulates feedback-guided revision of the
Task-Agent's own output under a limited budget. The final card
synthesizes how failure scope and intervention cost guide
component selection.

Applicability conditions determine when a procedure can be reused.
On a new task, Meta retrieves relevant revision guidance and
uses it to guide the Task-Agent in revising that task's own output from its feedback.
The Artifacts procedure and general principle express
methodological guidance rather than additional measured
intervention outcomes.

\paragraph{Harness skills: repair a reusable execution rule.}
The submission-contract skill applies when the required deliverable is source code but the agent returns prose or execution output. Its reusable content is the condition under which a complete-source check should be invoked and the execution behavior that enforces it. Reuse therefore changes how subsequent tasks are handled by the Harness. A valid submission can still contain an algorithmic error, so satisfying the output contract resolves an interface failure without establishing semantic correctness. This distinction explains why later feedback can motivate a different intervention route.

\paragraph{Model skills: construct an eligible learning signal.}
The replay skill describes how verified successful Task-Agent calls become training examples. Checkpoint provenance, task identity, native success, and target validity determine whether a call provides suitable evidence for the update. Reusing this skill means applying the same data-selection and training procedure to the current parent and available trajectories. R4 supplements M1 because it provides another observed response to that procedure. The added evidence informs when replay is useful; it does not imply that increasing replay volume always improves every task domain.

\paragraph{Artifacts skills: revise the current deliverable.}
The output-revision skill targets a defect that permitted feedback localizes within the Task-Agent's current submission. Its steps operate through the existing interface: inspect the feedback, edit the relevant output, test the revision, and resubmit within budget. The reusable element is this conditional repair procedure, including the circumstances in which further retries are unlikely to help. A later task supplies its own program or answer and its own feedback. Thus, the skill can guide a new revision without transferring the earlier answer or changing the Task-Harness.

\paragraph{General principles: connect diagnosis to intervention scope.}
A general principle connects these procedures through a shared decision criterion. The same failed coding task may require a submission check, learning from successful trajectories, or a local program edit. The relevant distinction is where the failure occurs and which available intervention can address it under the remaining budget. A bounded output repair is appropriate when the feedback identifies an actionable defect; a recurring execution problem motivates a Harness change, while a persistent capability gap with eligible training data motivates a Model update. The principle guides this comparison while component-specific skills supply the concrete procedure.

\subsection{Integrating Skills into the Meta-Harness}
\label{app:meta-integration}
Let $\mathcal S_t$ and $\mathcal P_t$ denote the component-specific skills and general principles distilled from $e_t$. Maintained skills replace the previous active Skill content. Meta then revises the Action interfaces, retrieval calls, stage-specific instructions, and supplementary prompts that consume those skills:
\begin{equation}
\begin{aligned}
\mathcal K_{t+1}&=\operatorname{Maintain}(\mathcal K_t,\mathcal S_t,\mathcal P_t),\\
H_{t+1}^{\mathrm{Meta}}
&=\operatorname{Update}(H_t^{\mathrm{Meta}},\mathcal K_{t+1}),\\
M_{t+1}&=(\phi,H_{t+1}^{\mathrm{Meta}}).
\end{aligned}
\end{equation}
For example, a receipt-verification skill informs which evidence diagnosis requests, which checks the Harness branch generates, and which counterexamples evaluation inspects. Updating a record and updating its invocation flow are recorded separately. The result is a changed Meta-Harness, not just a larger collection of text.

\paragraph{Reuse and generalization.}
\label{app:skill-generalization}
HHMMA illustrates knowledge development: H1 supplies a contract-checking criterion for H2 at R2; M1 accumulates training response evidence at R4; and R5 formulates a bounded output-repair procedure for local code defects. R2-to-R3 shifts from execution checks to Model updates when semantic errors persist. The case illustrates reuse within an evolving system; it does not isolate the causal contribution of an individual skill.

\paragraph{From a retrieved record to an executable intervention.}
Retrieval provides Meta with a procedure; integration guides its use in the next attempt. Applicability informs diagnosis, procedures inform modification briefs, and preservation requirements constrain construction and validation. For example, a receipt-grounded skill links an unsupported completion claim to the evidence needed for assessment. The Harness change implements this check; paired evaluation determines whether it improves task outcomes. This chain connects the stored lesson to an intervention whose effects can be examined.

\paragraph{Using feedback to refine subsequent decisions.}
Intervention responses inform library maintenance.
Improvements support procedures under observed conditions;
regressions reveal preservation failures.
Failed candidates can inform diagnosis even when the parent
Task state is retained. Supplementation adds evidence;
revision updates inappropriate guidance.
Task acceptance and Meta learning thus respond differently,
while lessons remain linked to their source cases.

\section{Extended Related Work}
\label{app:extended-related-work}

\paragraph{Agent System Design and Adaptation.}
Recent work on LLM agents explores approaches to adapting execution structures, maintaining persistent memory, and coordinating information acquisition. HarnessForge~\citep{chen2026harnessforgejointharnesspolicy} jointly evolves execution harnesses and reasoning policies through failure-guided harness modification and harness-conditioned policy alignment. For persistent memory, All-Mem~\citep{lv2026allmemagenticlifelongmemory} separates online retrieval from offline topology consolidation, preserving evidence while supporting retrieval under bounded computational budgets. Typed links enable budgeted expansion from anchors to archived evidence. For breadth-oriented information gathering, moreover, A-MapReduce~\citep{chen2026amapreduceexecutingwidesearch} decomposes wide-search objectives into parallel retrieval tasks and aggregates results, using experiential memory to refine task allocation and recomposition. These approaches improve how agents organize execution and retain or acquire information, with emphases on system compatibility, memory maintenance, and scalability.

\paragraph{Structured Supervision and Credit Assignment.}
A complementary line of research studies the structure and granularity of supervision for agent learning and language model post-training. GEAR~\citep{lv2026mitigatingfalsecreditpropagation} models prerequisite and activation dependencies among rubric criteria with a probabilistic graphical framework, suppressing unsupported criterion contributions when constructing scalar rewards. The resulting reward captures criterion dependencies at the response level. For long-horizon agent distillation, AlignOPSD~\citep{chen2026timestampsdecisionalignedonpolicydistillation} aligns privileged supervision with functionally corresponding decisions and distributes outcome-based credit across variable-duration decision spans and their constituent turns. TRCA~\citep{zhang2026trcatransitionwiserubriccredit} derives step-level supervision from action-induced state transitions using Evidence, Execution, and Invalidity rubrics. Its foundational reward evaluates local transition quality, while its breakthrough reward recognizes newly satisfied task conditions, providing fine-grained learning signals without requiring successful reference trajectories or learned process evaluators, even when terminal success feedback is sparse.

\section{Skill Records and Prompt Protocols}
\label{app:visual-templates}
\subsection{Overview}
\label{app:skill-illustrations}
\paragraph{Harness} (Fig.~\ref{fig:app-skill-record})
enforces complete-source submission, preserving unrelated behavior.

\paragraph{Model} (Fig.~\ref{fig:app-skill-model})
trains on successful calls,
checking provenance, success, and target validity.

\paragraph{Artifacts} (Fig.~\ref{fig:app-skill-artifacts})
guides code editing, testing, and resubmission
based on execution feedback.
Outputs remain task-bound; guidance is reusable.

\paragraph{General principle}
(Fig.~\ref{fig:app-skill-principle})
matches interventions to failure and budget,
guiding component-specific procedures
across different task scenarios.
\subsection{Prompt Templates}
\label{app:prompt-overview}

\paragraph{Selection} (Prompt~\ref{prompt:selection})
selects components using feedback and retrieved skill records,
recording expected effects and constraints.

\paragraph{Localization} (Prompt~\ref{prompt:localization})
distinguishes Harness weaknesses from task-specific symptoms.

\paragraph{Directions} (Prompt~\ref{prompt:directions})
defines a brief with interface and preservation constraints.

\paragraph{Generation} (Prompt~\ref{prompt:generation})
constructs a Harness candidate,
linking modules to its mechanism.

\paragraph{Model update} (Prompt~\ref{prompt:model})
specifies SFT data, training settings,
and checkpoint verification.

\paragraph{Output revision}
(Prompt~\ref{prompt:output_revision})
guides revision of Task-Agent outputs
using permitted feedback within budget,
keeping Model and Harness fixed.

\paragraph{Distillation} (Prompt~\ref{prompt:distillation})
distills outcomes into guidance for future component selection,
recording applicability, actions, and limitations.

\paragraph{Maintenance} (Prompt~\ref{prompt:maintenance})
retrieves, adds, supplements, revises, and merges records,
retaining evidence and cases for subsequent skill reuse.

\subsection{Protocol Cards}
\label{app:prompts}

\definecolor{rsiPromptGreen}{HTML}{289754}
\definecolor{rsiSkillGold}{HTML}{E8BE35}
\definecolor{rsiCardInk}{HTML}{161C21}
\tcbset{rsiProtocol/.style={
  enhanced jigsaw, breakable, sharp corners,
  colback=white, coltext=rsiCardInk,
  boxrule=0.8pt, boxsep=0pt,
  left=9pt, right=9pt, top=8pt, bottom=8pt,
  toptitle=5pt, bottomtitle=5pt,
  fonttitle={\normalfont\bfseries\fontsize{10.5}{12}\selectfont\hyphenpenalty=10000\relax\exhyphenpenalty=10000\relax},
  fontupper=\normalfont\ttfamily\fontsize{9.2}{11.6}\selectfont,
  halign title=left,
  before skip=9pt, after skip=9pt,
  pad at break*=3pt,
  lines before break=5,
  before upper={\raggedright\setlength{\parindent}{0pt}%
                \setlength{\parskip}{0pt}\hyphenpenalty=10000%
                \exhyphenpenalty=10000\emergencystretch=1em}
}}

\begingroup
\refstepcounter{figure}\label{fig:app-skill-record}
\begin{tcolorbox}[rsiProtocol,colframe=rsiSkillGold,colbacktitle=rsiSkillGold,
coltitle=black,title={Figure~\thefigure. Harness: Submission Contract},
title after break={Figure~\thefigure. Harness: Submission Contract (continued)}]
Record: typed\_\allowbreak{}program\_\allowbreak{}commit.v1\par
\smallskip
Scope: Harness\par
\smallskip
When: The public contract requires source code, but final submissions contain stdout or prose.\par
\smallskip
Steps:\par\nobreak
1. Localize submission-shape failures.\par
2. Gate complete-source instructions on the public program contract.\par
3. Keep shape retries inside the original loop and budget.\par
4. Test program entry and non-program bypass; inspect paired regressions.\par
\smallskip
Preserve: Model, Planning, Memory, tools, budget, and grader.\par
\smallskip
Avoid when: No public program contract exists, or the defect is algorithmic.\par
\smallskip
Evidence: R1: 106 -> 148 successes / 600; 54 gains and 12 regressions. Code: 31 gains, 5 regressions.\par
\smallskip
Status: One paired training batch. Full preservation was not achieved.\par
\smallskip
Reuse: Referenced by the R2 stateful-completion intervention.\par
\end{tcolorbox}
\endgroup
\begingroup
\refstepcounter{figure}\label{fig:app-skill-model}
\begin{tcolorbox}[rsiProtocol,colframe=rsiSkillGold,colbacktitle=rsiSkillGold,
coltitle=black,title={Figure~\thefigure. Model: Successful-Trajectory Replay},
title after break={Figure~\thefigure. Model: Successful-Trajectory Replay (continued)}]
Record: verified\_\allowbreak{}parent\_\allowbreak{}replay.v1\par
\smallskip
Scope: Model\par
\smallskip
When: Harness contracts are valid, success remains low, and verified successful parent rollouts are available.\par
\smallskip
Steps:\par\nobreak
1. Bind calls to checkpoint, task, and native success.\par
2. Exclude reporting tasks, Meta-edited answers, and unverified traces.\par
3. Filter malformed or overlength targets; train one-epoch LoRA.\par
4. Verify sample coverage and serving bindings; compare the full paired batch.\par
\smallskip
Preserve: Harness, artifacts, graders, budgets, memory input, and length filters.\par
\smallskip
Avoid when: Provenance is missing or extra data repeats coverage without addressing the diagnosed failure.\par
\smallskip
Evidence: R3: 3,798 calls; 134 -> 141 / 600 (34 gains, 27 regressions).\par
\smallskip
Reuse: R4: 5,397 calls; 159 -> 160 / 600 (23 gains, 22 regressions). The same procedure receives new evidence.\par
\end{tcolorbox}
\endgroup
\begingroup
\refstepcounter{figure}\label{fig:app-skill-artifacts}
\begin{tcolorbox}[rsiProtocol,colframe=rsiSkillGold,colbacktitle=rsiSkillGold,
coltitle=black,title={Figure~\thefigure. Artifacts: Feedback-Guided Revision},
title after break={Figure~\thefigure. Artifacts: Feedback-Guided Revision (continued)}]
Type: Reformulated procedure\par
\smallskip
Scope: Task outputs (code / answers)\par
\smallskip
When: Execution feedback identifies a repairable defect in the current code, and the remaining budget permits a short revision cycle.\par
\smallskip
Steps:\par\nobreak
1. Inspect permitted tests, stderr, or execution feedback.\par
2. Have the Task-Agent revise its own code to address the defect.\par
3. Retest and resubmit through the existing interface within budget.\par
4. Record the repair, outcome, regressions, and edit/test cost.\par
\smallskip
Preserve: Model parameters, Harness, task contract, evaluator, and execution limits.\par
\smallskip
Rationale: Local revision can improve the current submission without paying the cost of model training or Harness reconstruction.\par
\smallskip
Carry over: Meta distills repair conditions, steps, and observed limits into its skill library. The task-specific code or answer is not transferred to new tasks.\par
\smallskip
New task: Meta retrieves the skill to diagnose a new task and guide revision of that task's own output using its own feedback.\par
\smallskip
Avoid when: The defect requires a broader execution or capability change, or further local retries would exhaust the budget.\par
\end{tcolorbox}
\endgroup
\begingroup
\refstepcounter{figure}\label{fig:app-skill-principle}
\begin{tcolorbox}[rsiProtocol,colframe=rsiSkillGold,colbacktitle=rsiSkillGold,
coltitle=black,title={Figure~\thefigure. Principle: Scope and Budget},
title after break={Figure~\thefigure. Principle: Scope and Budget (continued)}]
Type: Synthesized principle\par
\smallskip
Scope: Component selection\par
\smallskip
When: Feedback exposes a bottleneck and the remaining budget constrains the next intervention.\par
\smallskip
Principle: Match intervention scope to the diagnosed failure. Prefer a bounded output revision when it can address the current defect; broader updates need evidence of a reusable bottleneck.\par
\smallskip
Steps:\par\nobreak
1. Local output defect with actionable feedback: revise and resubmit Artifacts.\par
2. Recurring execution or interface failure: consider a Harness repair.\par
3. Recurring capability gap with suitable training data: consider a Model update.\par
4. Weigh expected benefit and reuse against implementation, training, and validation cost.\par
\smallskip
Preserve: Task requirements, evaluation rules, and the available budget.\par
\smallskip
Avoid when: The cheaper action cannot address the failure, or repeated local repairs consume more budget than a reusable fix.\par
\smallskip
Verify: Use observed quality and cost to revise the next decision; treat the selected route as a hypothesis.\par
\end{tcolorbox}
\endgroup
\begingroup
\refstepcounter{rsiPrompt}\label{prompt:selection}\label{fig:prompt-selection}
\begin{tcolorbox}[rsiProtocol,colframe=rsiPromptGreen,colbacktitle=rsiPromptGreen,
coltitle=white,title={Prompt~\thersiPrompt. Component Selection (Partial)},
title after break={Prompt~\thersiPrompt. Component Selection (Partial) (continued)}]
Choose exactly one currently executable Task component: HARNESS, MODEL, or ARTIFACTS. Record the intervention proposal and its prospective expectations. Do not implement the intervention or update G/K during routing.\par
\smallskip
[...]\par
\smallskip
Explain the observed problem, a plausible mechanism, why this component is worth prioritizing now, the exact proposed change, and what subsequent evidence would support or contradict the proposal. A justified feasible choice is required; proving that every alternative is worse is not.\par
\smallskip
[...]\par
\smallskip
- rationale contains concise mechanism reasoning, relevant principles, counter-evidence, the three-component comparison, and limitations.\par
\smallskip
[...]\par
\smallskip
- Include expectations.implementation\_\allowbreak{}events, preserve, risks, unknowns, evidence, and principle\_\allowbreak{}ids. used\_\allowbreak{}principle\_\allowbreak{}ids names active principles actually used.\par
\end{tcolorbox}
\endgroup
\phantomsection\label{app:prompt-interventions}
\begingroup
\refstepcounter{rsiPrompt}\label{prompt:localization}\label{fig:prompt-localization}
\begin{tcolorbox}[rsiProtocol,colframe=rsiPromptGreen,colbacktitle=rsiPromptGreen,
coltitle=white,title={Prompt~\thersiPrompt. Harness Fault Localization (Partial)},
title after break={Prompt~\thersiPrompt. Harness Fault Localization (Partial) (continued)}]
\# Harness Failure Localization Agent\par
\smallskip
Your goal is to analyze the execution failures of the current winner harness \{winner\_\allowbreak{}harness\_\allowbreak{}name\} and produce a module-localized, evidence-grounded, generation-ready diagnosis of transferable harness weaknesses.\par
\smallskip
[...]\par
\smallskip
Use failed trajectories as evidence for diagnosing the harness, not as specifications for task-specific fixes.\par
\smallskip
[...]\par
\smallskip
For every major diagnosis, separate:\par
- the task-specific surface manifestation observed in the trajectory\par
- the deeper harness behavior that caused it\par
- why that behavior is expected to recur across different task families\par
- what module-level capability is missing or underdeveloped\par
\end{tcolorbox}
\endgroup
\begingroup
\refstepcounter{rsiPrompt}\label{prompt:directions}\label{fig:prompt-directions}
\begin{tcolorbox}[rsiProtocol,colframe=rsiPromptGreen,colbacktitle=rsiPromptGreen,
coltitle=white,title={Prompt~\thersiPrompt. Improvement Directions (Partial)},
title after break={Prompt~\thersiPrompt. Improvement Directions (Partial) (continued)}]
This phase does not generate code.\par
\smallskip
[...]\par
\smallskip
Do not recommend:\par
- replacing the whole benchmark loop\par
- changing the dataset or evaluator\par
- adding neural retraining\par
[...]\par
- hard-coding benchmark item IDs, answers, tool traces, or golden values\par
\end{tcolorbox}
\endgroup
\begingroup
\refstepcounter{rsiPrompt}\label{prompt:generation}\label{fig:prompt-generation}
\begin{tcolorbox}[rsiProtocol,colframe=rsiPromptGreen,colbacktitle=rsiPromptGreen,
coltitle=white,title={Prompt~\thersiPrompt. Harness Generation (Partial)},
title after break={Prompt~\thersiPrompt. Harness Generation (Partial) (continued)}]
Prefer the smallest coherent harness that implements the brief.\par
\smallskip
[...]\par
\smallskip
The candidate must be an independent bundle, not a patch against the winner or a dependent extension of another generated harness.\par
\smallskip
[...]\par
\smallskip
The Module Change Map must name the chosen mechanism, not just the affected file.\par
\end{tcolorbox}
\endgroup
\begingroup
\refstepcounter{rsiPrompt}\label{prompt:model}\label{fig:prompt-model}
\begin{tcolorbox}[rsiProtocol,colframe=rsiPromptGreen,colbacktitle=rsiPromptGreen,
coltitle=white,title={Prompt~\thersiPrompt. Model Update (Partial)},
title after break={Prompt~\thersiPrompt. Model Update (Partial) (continued)}]
Explicitly record source run, checkpoint, task ID, output/evaluator hashes, deduplication and domain coverage; exclude the 744 reporting cohort, Meta-repaired answers and failed/unverified traces.\par
\smallskip
[...]\par
\smallskip
Original one-epoch LoRA hyperparameters and global batch 8 remain fixed.\par
\smallskip
[...]\par
\smallskip
verify rank receipts, candidate checkpoint and restored serving bindings.\par
\end{tcolorbox}
\endgroup
\begingroup
\refstepcounter{rsiPrompt}\label{prompt:output_revision}\label{fig:prompt-output_revision}
\begin{tcolorbox}[rsiProtocol,colframe=rsiPromptGreen,colbacktitle=rsiPromptGreen,
coltitle=white,title={Prompt~\thersiPrompt. Task-Output Revision Template},
title after break={Prompt~\thersiPrompt. Task-Output Revision Template (continued)}]
Role: You are the Task-Agent revising your own output.\par
\smallskip
Task: \{task\_\allowbreak{}specification\}\par
Your previous output: \{own\_\allowbreak{}output\}\par
Available feedback: \{permitted\_\allowbreak{}feedback\}\par
Diagnostic guidance: \{revision\_\allowbreak{}guidance\}\par
Remaining budget: \{remaining\_\allowbreak{}budget\}\par
\smallskip
Identify the defect supported by the supplied feedback. Revise your own program, answer, or structured output and resubmit it through the existing interface.\par
\smallskip
Preserve the required output format. Use only the permitted tools and feedback. Keep the model and Harness unchanged. Do not copy an answer from another task.\par
\smallskip
Return the revised submission in the task's original output format.\par
\end{tcolorbox}
\endgroup
\phantomsection\label{app:prompt-distillation}
\begingroup
\refstepcounter{rsiPrompt}\label{prompt:distillation}\label{fig:prompt-distillation}
\begin{tcolorbox}[rsiProtocol,colframe=rsiPromptGreen,colbacktitle=rsiPromptGreen,
coltitle=white,title={Prompt~\thersiPrompt. Meta Feedback and Skill Formation (Partial)},
title after break={Prompt~\thersiPrompt. Meta Feedback and Skill Formation (Partial) (continued)}]
Turn what was expected, changed, and observed into conditional guidance for future component selection and implementation, and revise G when justified. Principle-memory changes, substantive G changes, later actual use, and performance improvement are distinct outcomes.\par
\smallskip
[...]\par
\smallskip
2. Stage 3 -- Derive principles. Retrieve related experiences, active principles, and counter-evidence. Specify conditions, recommended checks/actions, support, limitations, and invalidation criteria. Local implementation success and terminal failure can support differently scoped lessons. Do not mechanically filter by score sign or infer component exhaustion from one non-improving intervention.\par
\end{tcolorbox}
\endgroup
\phantomsection\label{app:prompt-maintenance}
\begingroup
\refstepcounter{rsiPrompt}\label{prompt:maintenance}\label{fig:prompt-maintenance}
\begin{tcolorbox}[rsiProtocol,colframe=rsiPromptGreen,colbacktitle=rsiPromptGreen,
coltitle=white,title={Prompt~\thersiPrompt. Skill Retrieval and Maintenance (Partial)},
title after break={Prompt~\thersiPrompt. Skill Retrieval and Maintenance (Partial) (continued)}]
These are advisory matches, not component rankings. Inspect avoid\_\allowbreak{}when, confidence, counterevidence and unknowns; use read\_\allowbreak{}skill for full records and original rollout references when needed. Empty retrieval is valid.\par
\smallskip
[...]\par
\smallskip
Meta decides semantic coverage and maintenance, not a similarity threshold. Inspect retrieved versions before creating near-duplicates.\par
\smallskip
Methods use kind=procedure; cases remain immutable observations. Supplementation does not automatically strengthen confidence.\par
\smallskip
[...]\par
\smallskip
One intervention is one support event; summaries or later uses of its ID are not independent replications.\par
\end{tcolorbox}
\endgroup
\par
\endgroup

\end{document}